%% file: acl.tex
\pdfoutput=1
\documentclass[11pt]{article}

\usepackage{acl}

\usepackage{times}
\usepackage{latexsym}
\usepackage{color}

\usepackage[T1]{fontenc}
\usepackage[utf8]{inputenc}

\usepackage{microtype}

\usepackage{booktabs}
\usepackage{todonotes}
\usepackage{multirow}
\usepackage{enumitem,kantlipsum}
\usepackage{siunitx}
\usepackage{chngpage}
\usepackage{amssymb}
\usepackage{changes}
\usepackage{mathtools}
\usepackage{enumitem}

\usepackage{subcaption}
\usepackage{multirow}

\usepackage{rotating}

\usepackage{longtable}
\usepackage{minitoc}
\usepackage{caption}
\usepackage{footnote}
\newcommand{\result}[2]{ #1 \color{lightgray}{\scriptsize{$\pm{#2}$}}}

\newcommand{\ignore}[1]{}

\title{\textit{Of Human Criteria and Automatic Metrics}:\\A Benchmark of the Evaluation of Story Generation}
\author{
    Cyril Chhun$^1$ \qquad Pierre Colombo$^{2}$\thanks{\ Previously from Laboratoire des Signaux et Systèmes (L2S), CentraleSupélec, CNRS, Université Paris-Saclay.} \qquad Fabian M. Suchanek$^1$ \qquad Chloé Clavel$^1$ \\
    $^1$LTCI, Télécom Paris, Institut Polytechnique de Paris \\
    $^2$Lab of Mathematics and Informatics (MICS), CentraleSupélec, Université Paris-Saclay\\
    \texttt{cyril.chhun@telecom-paris.fr}\\
}

\def\NAME{{\texttt{HANNA}}}
\begin{document}
\doparttoc % Tell to minitoc to generate a toc for the parts
\faketableofcontents % Run a fake tableofcontents command for the partocs

% \part{} % Start the document part
% \parttoc % Insert the document TOC

\maketitle
\begin{abstract}

Research on Automatic Story Generation (ASG) relies heavily on human and automatic evaluation. However, there is no consensus on which human evaluation criteria to use, and no analysis of how well automatic criteria correlate with them. In this paper, we propose to re-evaluate ASG evaluation. We introduce a set of 6 orthogonal and comprehensive human criteria, carefully motivated by the social sciences literature. We also present {\NAME}, an annotated dataset of 1,056 stories produced by 10 different ASG systems. {\NAME} allows us to quantitatively evaluate the correlations of 72 automatic metrics with human criteria. Our analysis highlights the weaknesses of current metrics for ASG and allows us to formulate practical recommendations for ASG evaluation.% and to identify future ASG research directions. %Corroborating previous work in other fields, we find that currently used automatic metrics (\textit{e.g.}\ \texttt{BLEU}) correlate poorly with human judgment. Finally, we formulate recommended practices for ASG evaluation and identify future ASG research directions.

\end{abstract}

% \listoftodos
% \todo[inline]{uncomment line 131 of \texttt{acl.sty}}

\section{Introduction}

Storytelling is at the heart of human societies: skillful storytelling allows a narrator to connect more authentically with their audience and listeners, and to understand the essence of complex concepts better \citep{suzuki2018dialogues}. Numerous applications could benefit from strong automatic story generation systems, including gaming \citep{hartsook2011toward}, communication \citep{alhussain2021automatic}, and education \citep{aylett2007fearnot}. Several approaches have been explored to generate stories automatically or with minimum editing efforts \citep{alabdulkarim2021automatic}. \emph{Automatic story generation} (ASG) takes as input a short sentence (a \emph{prompt}) and aims at generating a narrative from it \citep{cavazza2006narratology,lebowitz1985story}. Advances in neural language models \citep{radford2018improving, radford2019language, brown2020language} have allowed substantial progress in ASG. %, making generated stories harder to distinguish from human-made stories \citep{clark2021all}.
\\\noindent To further improve the quality of generated stories, it is indispensable to systematically evaluate ASG models. However, there is little work that specifically studies ASG evaluation. Most research works rely on human criteria such as coherence \citep{xu-etal-2018-skeleton,colombo2019affect,jalalzai2020heavy}, relevance  \citep{jhamtani-berg-kirkpatrick-2020-narrative}, overall quality \citep{brahman2020modeling}, narrative flow \citep{rashkin-etal-2020-plotmachines}, and creativity \citep{pascual-etal-2021-plug-play}. However, taken individually, these criteria fail to encompass all aspects of the task, and there is no consensus on a set of criteria that would cover those aspects in a complete and non-redundant fashion. Due to the high cost of human annotation, system quality is also often evaluated using automatic metrics. However, it is not clear how these metrics correlate with human judgment in ASG, and thus how suitable they are at all for the evaluation of ASG.\\
\noindent \textbf{Contributions}. In this work, we revisit both human and automatic evaluation of ASG. We believe that this meta-evaluation is a missing piece in the ASG literature and a crucial step to strengthening the foundations of ASG.
Formally, our contributions to the ASG field are:
\begin{enumerate}[wide, labelindent=0pt]
  \item \textbf{A comprehensive set of non-redundant human criteria for ASG evaluation.} Motivated by the social sciences literature \citep{mccabe1984makes, dickman2003four, bae2021preliminary}, we introduce six human criteria: relevance, coherence, empathy, surprise, engagement and complexity.
  \item \textbf{{\NAME}\footnote{The \NAME\ dataset and corresponding code are available on \url{https://github.com/dig-team/hanna-benchmark-asg}.}, a large annotated dataset of \underline{H}uman-\underline{AN}notated \underline{NA}rratives for ASG evaluation}, which contains 1,056 stories generated from 96 prompts. Each prompt is linked to a human story and stories generated by 10 different ASG generation systems. Each story was annotated by 3 different human raters along our 6 proposed human criteria.
  \item  \textbf{A meta-evaluation of ASG with fine-grained recommendations.} Relying on {\NAME}, we perform an extensive study of the performance of the ASG systems and we analyze the correlations of 72 existing automatic metrics with our proposed human criteria. The obtained results demonstrate the limitations of current automatic evaluation methods and allow us to make recommendations on which metrics to use for ASG evaluation.
\end{enumerate}
\section{Related work}

\subsection{Human evaluation}
\label{par:human_eval}

\citet{van-der-lee-etal-2019-best} advise to define separate and precise criteria for human evaluation to make it as accurate as possible.
However, in ASG, there is no consensus on the criteria to be used: among others, we find a pairing task \citep{fan2018hierarchical}, fluency and coherence \citep{xu-etal-2018-skeleton}, creativity \citep{pascual-etal-2021-plug-play}, faithfulness \citep{peng-etal-2018-towards, wang2020narrative}, fidelity \citep{yao2019plan}, grammar and logicality \citep{guan2019story, guan-etal-2020-knowledge}, overall quality and relevance \citep{jhamtani-berg-kirkpatrick-2020-narrative, goldfarb-tarrant-etal-2020-content, guan2021openmeva}, outline utilisation and narrative flow \citep{rashkin-etal-2020-plotmachines}, emotion faithfulness \citep{witon2018disney}, and content quality \citep{brahman2020modeling}. Many of these criteria are not specific to ASG (fluency, grammar, overall quality, content quality), overlap with one another (pairing task, faithfulness, and fidelity are variations of relevance; logicality and narrative flow, of coherence) or are ascribed to a specific setting (outline utilisation, emotion faithfulness). Furthermore, evaluation protocols mostly use only two or three criteria, which is not enough to grasp all aspects of a task as complex as ASG. 
They also do not associate Likert scales with explicit descriptions, even though such descriptions could reduce the subjectivity of the labelling process.

\subsection{Automatic evaluation}
\label{par:automatic_evaluation}
Although most of the research work in ASG relies on \texttt{BLEU} and \texttt{ROUGE}, there exists a plethora of automatic metrics to evaluate ASG. These can be classified into two categories: \emph{reference-based} ($\Xi$) metrics evaluate a candidate text by comparing it to a reference text (in our case, the human story), and \emph{reference-free} (¤) metrics rely only on the candidate story (and, possibly, on the prompt). In both categories, we find \emph{string-based} (§), \emph{embedding-based} ($\varepsilon$) and \emph{model-based} ($\Delta$) metrics. String-based metrics evaluate the textual representation of the inputs; they cannot handle synonyms or paraphrases. By contrast, embedding-based metrics rely on word embeddings, \textit{e.g.}\ word2vec \citep{mikolov2013efficient, mikolov2013distributed}, or contextualized embeddings, \textit{e.g.}\ obtained from BERT \citep{devlin-etal-2019-bert}. Finally, model-based metrics leverage regression or pre-trained language models to return a score. A synoptic classification can be found in \autoref{tab:metrics}\footnote{\label{foot:bartscore}\texttt{BARTScore} was designed to be either reference-based or reference-free depending on the setting.}.

\begin{table}[h]
\tiny\centering
\begin{tabular}{@{}ll@{}l@{}}
\toprule
  & Reference-based ($\Xi$) & Reference-free (¤) \\ 
\midrule
\multirow{6}{0.4cm}{String-based (§)} & \texttt{BLEU} \citep{papineni2002bleu} & \texttt{Coverage} \citep{grusky-etal-2018-newsroom}\\
& \texttt{ROUGE} \citep{lin2004rouge} & \texttt{Density} \citep{grusky-etal-2018-newsroom}\\
& \texttt{METEOR} \citep{banerjee2005meteor} & \texttt{Compression} \citep{grusky-etal-2018-newsroom}\\
& \texttt{\textsc{chrF}} \citep{popovic2015chrf} & \texttt{Text length} \citep{fabbri2021summeval}\\
& \texttt{CIDEr} \citep{vedantam2015cider} & \texttt{Novelty} \citep{fabbri2021summeval}\\
& & \texttt{Repetition} \citep{fabbri2021summeval}\\
\midrule
\multirow{5}{0.4cm}{Embed-ding-based ($\varepsilon$)} & \texttt{ROUGE-WE} \citep{ng2015better} & \\
& \texttt{BERTScore} \citep{zhang2019bertscore} & \\
& \texttt{MoverScore} \citep{zhao2019moverscore} & \texttt{SUPERT} \citep{gao2020supert} \\
& \texttt{BaryScore} \citep{colombo2021automatic} & \\
& \texttt{DepthScore} \citep{staerman2021pseudo}\\
\midrule
\multirow{4}{0.4cm}{Model-based ($\Delta$)} & \texttt{S3} \citep{peyrard2017s3} & \\
& \texttt{SummaQA} \citep{scialom-etal-2019-answers} & \texttt{BLANC} \citep{vasilyev2020fill} \\
& \texttt{InfoLM} \citep{colombo2021infolm} \\
& \multicolumn{2}{c}{\texttt{BARTScore} \citep{yuan2021bartscore}}\\
\bottomrule
\end{tabular}
\caption{Classification of the automatic metrics considered in our study with symbols for easier identification.}
\label{tab:metrics}
\end{table}

\subsection{Meta-evaluation} Several previous works have studied the relationship between automatic metrics and human judgment \citep{zhang2004interpreting,ma-etal-2019-results}, reporting weak correlation \citep{novikova2017we,stent2005evaluating,mathur-etal-2020-tangled} and strong bias towards specific systems \citep{callison-burch-etal-2006-evaluating}.  Meta-evaluation has been done in image description \citep{elliott-keller-2014-comparing}, dialogue response generation \citep{liu-etal-2016-evaluate}, question generation \citep{nema-khapra-2018-towards}, table-to-text generation \citep{dhingra-etal-2019-handling}, question answering \citep{chen-etal-2019-evaluating}, and summarization \citep{bhandari2020re}. In ASG, \citet{guan2021openmeva} introduced the OpenMEVA benchmark which compares the overall quality of human and generated stories; their work especially focused on the textual features of stories. We build upon it and perform a comprehensive analysis of the correlations between 72 automatic metrics and 6 human criteria specifically tailored for ASG. 

\section{{\NAME} for ASG evaluation}

\subsection{ASG datasets}
% Story evaluation has been widely studied in different scenarii. \texttt{ROCStories} \citep{mostafazadeh2016corpus}, a corpus of 50k 5-sentence stories with titles, was designed for the Story Cloze Test: the prediction of the final sentence of a story given the four others. \citet{huang2016visual} developed the \texttt{VisualStorytelling} dataset, which contains sequences of images with corresponding descriptions divided in three tiers of temporal context. More recently, \citet{ammanabrolu20world} proposed the \texttt{WorldGeneration} dataset which adapts story generation to adventure games by guiding the generation process with location, character and object triplets.

% \noindent In this work, we instead choose to focus on the \texttt{WritingPrompts} (\texttt{WP}) dataset\footnote{We disregarded \texttt{ROCStories} because, while it is also used in several works, it focuses on the Story Cloze Test and the shortness of the stories made it less suited for story evaluation.}. An example prompt and story from \texttt{WP} can be found in \autoref{tab:example_prompt_story} \citep{fan2018hierarchical}, which contains stories generated from short sentences called \emph{prompts}, as it has been extensively used in previous literature for the design of ASG models \citep{rashkin-etal-2020-plotmachines, goldfarb-tarrant-etal-2020-content, fang2021transformer, wilmot2021temporal, guan2021long}.

Story evaluation has been widely studied in different scenarii. \texttt{ROCStories} \citep{mostafazadeh2016corpus}, a corpus of 50k 5-sentence stories with titles, was designed for the Story Cloze Test: the prediction of the final sentence of a story given the four others. \citet{huang2016visual} developed the \texttt{VisualStorytelling} dataset, which contains sequences of images with corresponding descriptions divided in three tiers of temporal context. More recently, \citet{ammanabrolu20world} proposed the \texttt{WorldGeneration} dataset which adapts story generation to adventure games by guiding the generation process with location, character and object triplets. The \texttt{WritingPrompts} (\texttt{WP}) dataset \citep{fan2018hierarchical} contains stories generated from short sentences called \emph{prompts}.
\noindent For our work, we chose the \texttt{WP} dataset, because it has been extensively used in previous literature for the design of ASG models \citep{rashkin-etal-2020-plotmachines, goldfarb-tarrant-etal-2020-content, fang2021transformer, wilmot2021temporal, guan2021long}. While \texttt{ROCStories} has also been used in several works, the shortness of the stories made it less suited for our evaluation. An example prompt and story from \texttt{WP} is shown in  \autoref{tab:example_prompt_story} \citep{fan2018hierarchical}.

\subsection{Chosen setting}
\label{ssec:setting}
{\NAME}, our annotated dataset for ASG, contains outputs from 10 different systems aligned on 96 common prompts with human stories from the \texttt{WP} dataset, for 1,056 stories in total, with 3 human annotations per story (19,008 annotations in total) and automatic metric scores, allowing for an analysis of the correlations between these metrics (\autoref{sec:analysis}).

\subsection{Chosen ASG sytems}
\label{ssec:asg_systems}
\noindent We directly contacted the authors of articles that introduced ASG systems and asked for the outputs of their systems. We managed to collect the outputs of \textbf{3 ASG systems}\footnote{We also collected outputs from two other systems \citep{goldfarb-tarrant-etal-2020-content, bai2020semantics}; unfortunately, these were not aligned with the others.} on the \texttt{WP} dataset: \texttt{Fusion} \citep{fan2018hierarchical}, \texttt{HINT} \citep{guan2021long}, and \texttt{TD-VAE} \citep{wilmot2021temporal}. We extracted 96 stories aligned on common prompts.
We then fine-tuned \textbf{7 pre-trained language models} for ASG on a causal language modeling task on \texttt{WP} to generate stories on the same 96 prompts, using the \emph{Transformers} library \citep{wolf-etal-2020-transformers}\footnote{\url{https://github.com/huggingface/transformers}}. We trained \texttt{BertGeneration} \citep{rothe2020leveraging}, \texttt{CTRL} \citep{keskar2019ctrl},    \texttt{RoBERTa} \citep{liu2019roberta}, \texttt{XLNet} \citep{yang2019xlnet}, \texttt{GPT} \citep{radford2018improving}, \texttt{GPT-2} \citep{radford2019language}, and \texttt{GPT-2 (tag)}, another instance of \texttt{GPT-2} trained with \texttt{<EOP>} (\underline{E}nd \underline{O}f \underline{P}rompt) tags, as inspired by \citet{bai2020semantics}, who argued that such tags could improve generation. %
% En général j'en cache dans le supplementary. 
\subsection{Proposed human criteria}\label{ssec:asg_human}
As mentioned in \autoref{par:human_eval}, there is no consensus on human criteria for ASG evaluation.
At the same time, work in social sciences has looked extensively at the features that make for a ``good'' story \citep{mccabe1984makes, dickman2003four, bae2021preliminary}. We condense them as follows into a new, comprehensive set of criteria:
\begin{enumerate}[wide, labelindent=0pt]
    \item  \textbf{Relevance} (\texttt{RE}): how well the story matches its prompt, used in \citet{jhamtani-berg-kirkpatrick-2020-narrative, goldfarb-tarrant-etal-2020-content};
    \item \textbf{Coherence} (\texttt{CH}): how much the story makes sense, used in \citet{xu-etal-2018-skeleton, peng-etal-2018-towards, yao2019plan, pascual-etal-2021-plug-play};
    \item  \textbf{Empathy} (\texttt{EM}): how well the reader understood the character’s emotions, derived from the importance of emotional commentary \citep{mccabe1984makes}, passion \citep{dickman2003four}, and empathy \citep{keen2007empathy, bae2021preliminary};
    \item  \textbf{Surprise} (\texttt{SU}): how surprising the end of the story was, derived from the importance of schema violation, or unexpectedness \citep{schank1978interestingness, bae2021preliminary}, postdictability \citep{behrooz2019story}, and novelty \citep{randall1999narrative};
    \item  \textbf{Engagement} (\texttt{EG}): how much the reader engaged with the story; a more subjective criterion associated with projecting volitive modality (making the reader formulate a subjective judgment and express a desire to see something accomplished) \citep{toolan2012engagement} and story outcome, which is an underlying cause of story liking \citep{iran1987cognitive}; 
    \item \textbf{Complexity} (\texttt{CX}): how elaborate the story is; derived from the importance of detailed descriptions and sophisticated problem-solving \citep{mccabe1984makes} and good world-building \citep{roine2016imaginative}.
\end{enumerate}
The four last criteria are an original contribution and were designed to evaluate story features that are different from the first two criteria (\texttt{RE} and \texttt{CH}), which are currently most used in the ASG literature. Examples of annotations w.r.t.\ those criteria are shown in \autoref{tab:example_prompt_story}.
\begin{table*}[h!]
\small\centering
\begin{minipage}{0.4\textwidth}
\noindent \textbf{Prompt}: When you die, the afterlife is an arena where you face every insect and animal you killed in your life. If you win you go to heaven, lose you go to hell. Your job was an exterminator on earth.\\
\\
\noindent \textbf{Human}: 3,000 years have I been fighting. Every morning, the raccoons scratch at my eyes. Every evening, the skunks spray me while the opossums chew at my feet. I have never had any tools. I have only my hands. I don’t remember the place I came from before this. [...]\\
\\
\noindent \textcolor{blue}{\textbf{Story \#1}}: First of all, not everyone was entitled to be an exterminator. But the ones that were – maybe were, like, \emph{genius}, because, yes, I had once belonged to a less fortunate class of people – had all the opportunity to work for the damn plant killer, and it's hard work. [...]\\
\\
\noindent \textcolor{violet}{\textbf{Story \#2}}: It was hell. Not exactly a place of torture. There were no guards in prison and you couldn't just walk through it, either, because you would get killed regardless. hell was a young man, and he was lying on his floor. He was unconscious. [...]\\
\end{minipage}
\hspace{0.5cm}
\begin{minipage}{0.5\textwidth}
\centering
\begin{tabular}{cccccccc}
\toprule
Story & \texttt{RE} & \texttt{CH} & \texttt{EM} & \texttt{SU} & \texttt{EG} & \texttt{CX}\\
\midrule
\multirow{3}{*}{Human} & 5 & 5 & 1 & 3 & 4 & 1 \\
& 2 & 2 & 3 & 2 & 2 & 3\\
& 4 & 4 & 3 & 2 & 4 & 4 \\
\midrule
\multirow{3}{*}{\color{blue}Story \#1} & 2 & 4 & 3 & 1 & 1 & 1 \\
& 2 & 2 & 2 & 1 & 2 & 2\\
& 2 & 3 & 2 & 3 & 3 & 3 \\
\midrule
\multirow{3}{*}{\color{violet}Story \#2} & 5 & 5 & 3 & 3 & 3 & 2 \\
& 3 & 2 & 3 & 2 & 2 & 3\\
& 3 & 4 & 3 & 4 & 4 & 3 \\
\bottomrule
\end{tabular}
\begin{tabular}{lrrr}
\\
\toprule
    \multicolumn{1}{c}{Metric} & \text{Human} & \text{\color{blue}Story \#1} & \text{\color{violet}Story \#2} \\
    \midrule
    \texttt{BLEU}$^{\Xi\text{§}}$ (\%) & 1.00 & 0.01 & 0.01 \\
    \texttt{ROUGE-1}$^{\Xi\text{§}}$ & 1.00 & 0.24 & 0.33 \\
    \texttt{chrF}$^{\Xi\text{§}}$ (\%) & 1.00 & 0.32 & 0.39 \\
    \texttt{BERTScore}$^{\Xi\varepsilon}$ & 1.00 & 0.50 & 0.52 \\
    \texttt{MoverScore}$^{\Xi\varepsilon}$ & 1.00 & 0.51 & 0.51 \\
    \texttt{BaryScore}$^{\Xi\varepsilon}$ & 0.00 & 0.92 & 0.92 \\
    % \texttt{DepthScore} & 0.00 & 0.12 & 0.11 \\
    \texttt{S3}$^{\Xi\Delta}$ & 1.39 & 0.07 & 0.15 \\
    \texttt{BARTScore}$^{\Xi\Delta}$ & -0.98 & -3.97 & -4.03 \\
    \texttt{SUPERT}$^{\text{¤}\varepsilon}$ & 0.94 & 0.37 & 0.36 \\
\bottomrule
\end{tabular}
\end{minipage}
\caption{Example prompt, human and generated stories from {\NAME} with human annotations and metric scores}
\label{tab:example_prompt_story}
\end{table*}

\subsection{Annotation Protocol}\label{ssec:annotation_proto}
\label{sub:amt_experiment}
To evaluate our human criteria on the 1,056 stories of {\NAME}, we conducted an annotation campaign on Amazon Mechanical Turk. 
As advised by \citet{karpinska2021perils}, for each task, we provided the human story alongside the story to be annotated, so that the workers could calibrate their judgment. 
Each of the stories was evaluated by three workers on our six proposed criteria. For this evaluation, we chose a 5-point Likert scale rather than a rank-based comparison because we reckoned that it would be tedious to order the large number of evaluated systems. We estimated that a HIT should take between 90 and 120 seconds, so we set the remuneration at \$0.28 per HIT, or between \$8.40 and \$11.40 per hour. To ensure that annotators spoke fluent English, we restricted access to the experiment to workers located in the UK, the US, Canada, Australia and New Zealand. We further required them to have the Masters Qualification. To remove noisy annotations and ensure that the workers read the stories, we chose to reject judgments that were made in fewer than 30 seconds. We additionally asked workers to write down the name of the first-mentioned fictional character of the story. The detailed instructions of the experiment and the inter-annotator agreement analysis can be found in the appendix (see \autoref{sec:amt_experiment_instructions} and \autoref{sub:inter_annotator_agreement}). Finally, following the recommendations of \citet{shapira-etal-2019-crowdsourcing}, we obtained the human score of a story by averaging the results of the three workers.

\subsection{Meta-evaluation strategies}
\label{ssec:meta_eval}
\noindent\textbf{Notations.} Let $y_i^j$ be the story generated by system $j \in \{1,\dots,S\}$ for prompt $i \in \{1,\dots,N\}$, and $m(y_i^j)$ the score associated to $y_i^j$ by a (human or automatic) metric $m$. Given a correlation coefficient $K$ (\textit{e.g.}\ Pearson's $r$ \citep{leusch2003novel}, Spearman's $\rho$ \citep{melamed2003precision} or Kendall's $\tau$ \citep{kendall1938new}), two meta-evaluation strategies are commonly used to evaluate metric quality.

\noindent\textbf{Story-level correlation ($K^\textrm{story}_{m_1,m_2}$)} measures how suited $m_1$ is w.r.t.\ $m_2$ if used as a loss or reward for a model. The correlation is applied to each story among all system outputs and the mean is taken. Formally:
\vspace*{-0.5\baselineskip}
\begin{align}
K^\textrm{story}_{m_1,m_2} & \triangleq  \frac{1}{N} \sum_{i=1}^N K ( \mathbf{C}^{\textrm{story}}_{m_1, i}, \mathbf{C}^{\textrm{story}}_{m_2, i} ),\\
\text{where} \quad \mathbf{C}^\textrm{story}_{m, i} & \triangleq \big{[}m(y_i^1),\cdots,m(y_i^S)\big{]}.\nonumber
\end{align}
\vspace*{-0.5\baselineskip}

\noindent\textbf{System-level correlation ($K^\textrm{sys}_{m_1,m_2}$)} measures how suited $m_1$ is w.r.t.\ $m_2$ if used to compare the performance of two systems. The correlation is applied to the mean values over all stories for all systems for both metrics. Formally:
\vspace*{-0.5\baselineskip}
\begin{align}
K^\textrm{sys}_{m_1,m_2} &\triangleq   K \left(\frac{1}{N} \mathbf{C}^\textrm{sys}_{m_1},\frac{1}{N} \mathbf{C}^\textrm{sys}_{m_2} \right),\\
\text{where} \quad \mathbf{C}^\textrm{sys}_{m} & \triangleq \left[ \sum\limits_{i=1}^N m(y_i^1),\dots, \sum\limits_{i=1}^N m(y_i^S)\right].\nonumber
\end{align}
\vspace*{-0.5\baselineskip}

\noindent\textbf{Statistical significance.} Correlations computed for two automatic metrics on the same annotated dataset are not independent. We follow \citet{graham-baldwin-2014-testing} and use the Williams test \citep{williams1959regression,moon2019williams}\footnote{\url{https://github.com/inmoonlight/nlp-williams}} to evaluate the significance of an increase in dependent correlations \citep{steiger1980tests}.

\section{{\NAME} Analysis}
\label{sec:analysis}
In this section, we analyse the scores of {\NAME} in detail. \autoref{tab:system_averages} shows that human stories achieve significantly higher scores than generated stories. Following \citet{mathur-etal-2020-tangled}, who advise to remove outliers, we compute correlations with human stories removed\footnote{The same applies for \autoref{sec:analysis_metrics}.}.

\subsection{Inter-annotator agreement}
To estimate the reliability of the annotations, we computed an intra-class coefficient for each criterion. Among the annotators which took part in the experiment, three of them covered 2490 stories, \textit{i.e.}, more than 78\% of the dataset, but no annotator graded the same story twice. Since the reliability is to be estimated for the average of the three ratings, the ICC2k coefficient (ICC for \emph{average random raters}) is the most relevant one, according to \citet{hallgren2012computing}. In particular, it accounts for the systematic errors of raters and random residual errors. The results are shown in \autoref{tab:icc}.

Coefficients are dispersed between 29\% and 56\% with relatively small confidence intervals (except for \texttt{RE} and \texttt{CH}), which can be considered between ``fair'' and ``moderate'' according to \citet{landis1977measurement}. These values are in tune with existing literature \citep{karpinska2021perils, habernal2017argumentation, spooren2010coding, ritter2011data, graham2017can} and show the difficulty of evaluating natural language generation. Therefore, we follow the methodology of \citet{craggs2005evaluating} and \citet{artstein2008inter}, who argue against setting a specific agreement threshold as long as there is a detailed reporting of the methodology (see \autoref{sub:amt_experiment} and \autoref{tab:amt_experiment_instructions}) and confidence intervals (\autoref{tab:icc}).  %The intrinsically subjective nature of ASG evaluation thus remains an open problem \citep{karpinska2021perils}.
%Thus, we reckon that our results are acceptable, given that the evaluation of a story will arguably always involve subjective judgment, no matter how accurately we define our annotation guidelines.

\label{sub:inter_annotator_agreement}
\begin{table}[!h]
\centering
\begin{tabular}{lccc}
\toprule
Criterion & \color{gray}{LB} & ICC2k & \color{gray}{UB} \\
\midrule
\texttt{RE} & \color{gray}{0.18} & 0.48 &  \color{gray}{0.65} \\
\texttt{CH} & \color{gray}{0.10} & 0.29 &  \color{gray}{0.48} \\
\texttt{EM} & \color{gray}{0.25} & 0.34 & \color{gray}{0.41} \\
\texttt{SU} & \color{gray}{0.16} & 0.28 & \color{gray}{0.37} \\
\texttt{EG} & \color{gray}{0.34} & 0.46 & \color{gray}{0.55} \\
\texttt{CX} & \color{gray}{0.48} & 0.56 & \color{gray}{0.63} \\
\bottomrule
\end{tabular}
\caption{Intra-class coefficient per criterion. LB and UB are the lower- and upper-bounds of the 95\% confidence interval}
\label{tab:icc}
\end{table}

\begin{table*}[!h]
\centering
\small
\begin{tabular}{lrrrrrrr}
\toprule
Model & \multicolumn{1}{c}{\texttt{RE}} & \multicolumn{1}{c}{\texttt{CH}} & \multicolumn{1}{c}{\texttt{EM}} & \multicolumn{1}{c}{\texttt{SU}} & \multicolumn{1}{c}{\texttt{EG}} & \multicolumn{1}{c}{\texttt{CX}} & \multicolumn{1}{c}{Average} \\
\midrule
Human          &  \result{4.17}{0.14} & \result{4.43}{0.10} &  \result{3.22}{0.14} &  \result{3.15}{0.15} &  \result{3.88}{0.12} &  \result{3.73}{0.13} & \result{3.76}{0.06} \\
\midrule
BertGeneration &  \result{2.46}{0.16} &  \result{3.14}{0.16} &  \result{2.28}{0.13} &  \result{\textbf{2.09}}{0.13} &  \result{2.67}{0.12} &  \result{2.41}{0.11} & \result{2.51}{0.06} \\
CTRL           &  \result{2.54}{0.16} &  \result{2.93}{0.16} &  \result{2.26}{0.13} &  \result{1.93}{0.12} &  \result{2.53}{0.12} &   \result{2.23}{0.10} & \result{2.40}{0.06} \\
GPT            &   \result{2.40}{0.16} &  \result{\textbf{3.22}}{0.15} &  \result{\textbf{2.37}}{0.12} &  \result{\textbf{2.13}}{0.13} &  \result{2.76}{0.13} &  \result{2.49}{0.12} & \result{2.56}{0.06} \\
GPT-2          &  *\result{\textbf{2.81}}{0.16} &  \result{\textbf{3.29}}{0.14} &  *\result{\textbf{2.47}}{0.12} & \result{\textbf{2.21}}{0.13} &  \result{\textbf{2.86}}{0.12} &   \result{2.68}{0.10} & \result{\textbf{2.72}}{0.06} \\
GPT-2 (tag)    &  \result{\textbf{2.67}}{0.16} &  *\result{\textbf{3.31}}{0.15} &  *\result{\textbf{2.47}}{0.12} &  *\result{\textbf{2.22}}{0.13} &  *\result{\textbf{2.92}}{0.12} &   *\result{\textbf{2.80}}{0.11} & *\result{\textbf{2.73}}{0.06} \\
RoBERTa        &  \result{2.54}{0.16} &  \result{3.22}{0.16} &  \result{2.27}{0.12} &  \result{\textbf{2.12}}{0.13} &  \result{2.74}{0.12} &  \result{2.41}{0.11} & \result{2.55}{0.06} \\
XLNet          &  \result{2.39}{0.17} &  \result{2.88}{0.16} &   \result{2.10}{0.12} &  \result{1.95}{0.12} &  \result{2.46}{0.13} &  \result{2.36}{0.11} & \result{2.36}{0.06} \\
Fusion         &  \result{2.09}{0.16} &  \result{2.86}{0.16} &  \result{1.99}{0.12} &  \result{1.72}{0.12} &  \result{2.27}{0.14} & \result{ 1.92}{0.11} & \result{2.14}{0.06} \\
HINT           &  \result{2.29}{0.16} &  \result{2.38}{0.16} &  \result{1.74}{0.13} &  \result{1.56}{0.11} &  \result{1.75}{0.12} &   \result{1.45}{0.10} & \result{1.86}{0.06}\\
TD-VAE         &  \result{2.51}{0.16} &  \result{2.99}{0.15} &  \result{2.07}{0.11} &   \result{\textbf{2.10}}{0.12} &  \result{2.59}{0.12} &  \result{2.49}{0.11} & \result{2.46}{0.06} \\
\bottomrule
\end{tabular}
\caption{Average system ratings per criterion with 95\% confidence interval. Best value in \textbf{bold} marked with an asterisk (*), values in the confidence interval of the best value in \textbf{bold} without asterisk}
\label{tab:system_averages}
\end{table*}

\subsection{Evaluating our human criteria}

\begin{figure}[h]
\centering
\begin{minipage}{0.45\columnwidth}
    \centering
    \includegraphics[width=\textwidth]{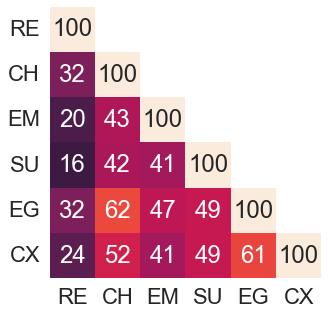}
    \caption{Story-level Kendall correlations (\%) between human criteria}
    \label{fig:story_level_human_correlations}
\end{minipage}
\hspace{0.2cm}
\begin{minipage}{0.45\columnwidth}
    \centering
    \includegraphics[width=\textwidth]{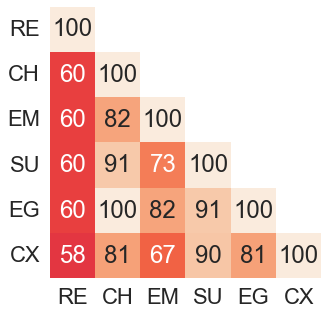}
    \caption{System-level Kendall correlations (\%) between human criteria}
    \label{fig:system_level_human_correlations}
\end{minipage}
\end{figure}
In this experiment, we study the relationship between the proposed human criteria. To compute story-level (\autoref{fig:story_level_human_correlations}) and system-level (\autoref{fig:system_level_human_correlations}) correlations, we average the human ratings.\\
\noindent \textbf{Story-level analysis (\autoref{fig:story_level_human_correlations})}. Kendall correlations range from 16\% (\texttt{RE}--\texttt{SU}) to 62\% (\texttt{CH}--\texttt{EG}), averaging at 40.7\%. In the appendix, we also show correlations with Spearman's $\rho$ (\autoref{fig:story_level_human_correlations_spearman}) and Pearson's $r$ (\autoref{fig:story_level_human_correlations_pearson}). \texttt{EG} correlates slightly more with \texttt{CH} and \texttt{CX}; this could confirm that coherent and intricate plots make readers more likely to connect with a story. In contrast, \texttt{RE} is poorly correlated to the other criteria, which makes sense: an excellent story in every other aspect can be completely unrelated to a prompt, and vice versa. Overall, moderate to weak correlations suggest that our criteria evaluate distinct aspects of storytelling which cannot be regrouped in fewer criteria.\\
\noindent \textbf{System-level analysis (\autoref{fig:system_level_human_correlations})}. Compared to story-level correlations, system level correlations are higher. Spearman (\autoref{fig:system_level_human_correlations_spearman}) and Pearson (\autoref{fig:system_level_human_correlations_pearson}) correlations are also higher than their story-level counterparts. This suggests that a given system tends to be uniformly better or worse than other systems across all criteria.

\subsection{Finding the best systems}
On \autoref{tab:system_averages}, we observe that generic fine-tuned models perform better than ASG systems according to human annotators. The best system is \texttt{GPT-2}, which scores better than all other systems on all criteria. The \texttt{GPT-2} variant trained with \texttt{<EOP>} tags shows marginal improvement compared to \texttt{GPT-2}, as reported in \citet{bai2020semantics}. It is worth noting that all models are still noticeably below human performance, which emphasizes that current systems are still a long way from human-like narrative intelligence.

\begin{figure*}[!ht]
\centering
\begin{minipage}{0.47\linewidth}
    \centering
    \includegraphics[width=\columnwidth]{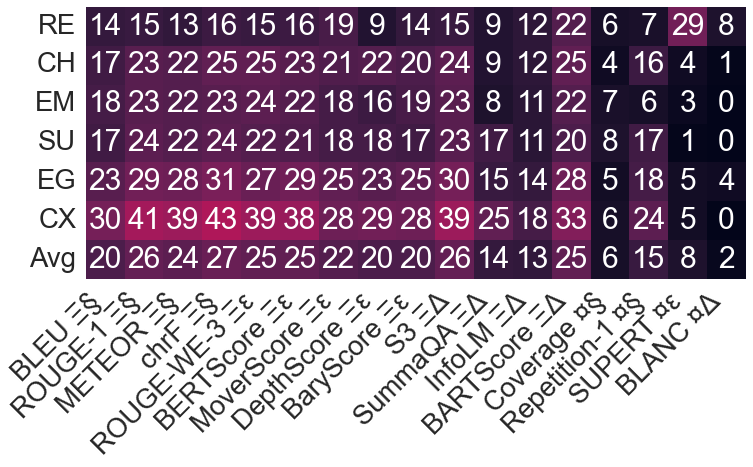}
    \caption{Story-level absolute Kendall correlations (\%) between metrics and criteria. Full version: \autoref{fig:story_level_mixed_correlations_kendall}.}
    \label{fig:story_level_mixed2_correlations_kendall}
\end{minipage}
\hspace{0.3cm}
\begin{minipage}{0.47\linewidth}
    \centering
    \includegraphics[width=\columnwidth]{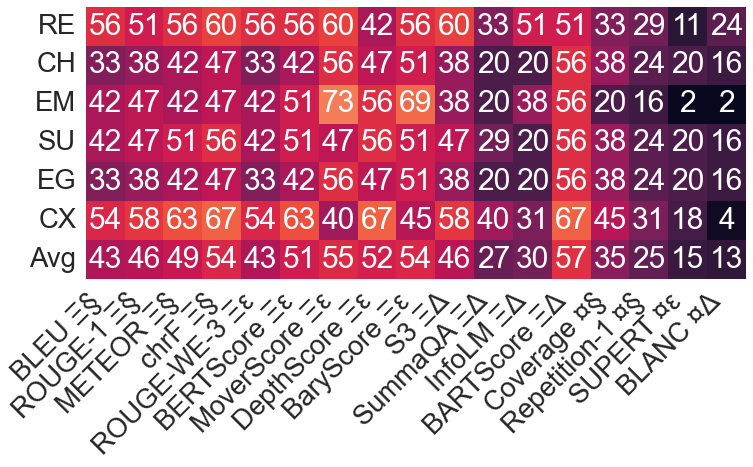}
    \caption{System-level absolute Kendall correlations (\%) between metrics and criteria. Full version: \autoref{fig:system_level_mixed_correlations_kendall}.}
    \label{fig:system_level_mixed2_correlations_kendall}
\end{minipage}
\end{figure*}

\begin{table*}[h]
\centering
\small
\begin{tabular}{c|cl@{}rl@{}rl@{}r}
\toprule
Level & Criterion & \multicolumn{1}{c}{Metric \#1} & \multicolumn{1}{c}{$|r|$ (\%)} & \multicolumn{1}{c}{Metric \#2} & \multicolumn{1}{c}{$|r|$ (\%)} & \multicolumn{1}{c}{Metric \#3} & \multicolumn{1}{c}{$|r|$ (\%)}\\
\midrule
\multirow{6}{*}{Story} & \texttt{RE} & \texttt{BARTScore}$^{\text{¤}\Delta}_2$ & 42.6 & \texttt{SUPERT}$^{\text{¤}\varepsilon}_1$ & 41.2 & \texttt{SUPERT}$^{\text{¤}\varepsilon}_2$ & 40.2 \\
\cline{2-8}
& \texttt{CH} & \texttt{Repetition-3}$^{\text{¤§}}$ & 38.1 & \texttt{BERTScore}$^{\Xi\varepsilon}_R$ & 37.1 & \texttt{S3}$^{\Xi\Delta}$ & 37.1 \\
\cline{2-8}
& \texttt{EM} & \texttt{S3}$^{\Xi\Delta}$ & 32.8 & \texttt{chrF}$^{\Xi\text{§}}$ & 32.4 & \texttt{BERTScore}$^{\Xi\varepsilon}_R$ & 32.1 \\
\cline{2-8}
& \texttt{SU} & \texttt{Novelty-1}$^{\text{¤§}}$ & 32.9 & \texttt{chrF}$^{\Xi\text{§}}$ & 32.7 & \texttt{ROUGE-1}$^{\Xi\text{§}}$ & 31.3 \\
\cline{2-8}
& \texttt{EG} & \texttt{BERTScore}$^{\Xi\varepsilon}_R$ & 43.0 & \texttt{Novelty-1}$^{\text{¤§}}$ & 42.3 & \texttt{chrF}$^{\Xi\text{§}}$ & 41.1 \\
\cline{2-8}
& \texttt{CX} & \texttt{chrF}$^{\Xi\text{§}}$ & 58.8 & \texttt{BERTScore}$^{\Xi\varepsilon}_R$ & 55.8 & \texttt{ROUGE-1}$^{\Xi\text{§}}$ & 55.0 \\
\midrule
\multirow{6}{*}{System} & \texttt{RE} & \texttt{ROUGE-S*}$^{\Xi\text{§}}_F$ & 80.4 & \texttt{ROUGE-SU*}$^{\Xi\text{§}}_F$ & 80.3 & \texttt{ROUGE-S*}$^{\Xi\text{§}}_R$ & 80.2 \\
\cline{2-8}
& \texttt{CH} & \texttt{BaryScore}$^{\Xi\varepsilon}_1$ & 88.2 & \texttt{BaryScore}$^{\Xi\varepsilon}_2$ & 88.0 & \texttt{BERTScore}$^{\Xi\varepsilon}_F$ & 87.9 \\
\cline{2-8}
& \texttt{EM} & \texttt{BaryScore}$^{\Xi\varepsilon}_1$ & 90.0 & \texttt{BaryScore}$^{\Xi\varepsilon}_2$ & 90.0 & \texttt{BERTScore}$^{\Xi\varepsilon}_F$ & 88.7 \\
\cline{2-8}
& \texttt{SU} & \texttt{BARTScore}$^{\Xi\Delta}_1$ & 92.7 & \texttt{BERTScore}$^{\Xi\varepsilon}_R$ & 91.1 & \texttt{DepthScore}$^{\Xi\varepsilon}$ & 90.7 \\
\cline{2-8}
& \texttt{EG} & \texttt{DepthScore}$^{\Xi\varepsilon}$ & 93.4 & \texttt{BARTScore}$^{\Xi\Delta}_1$ & 92.4 & \texttt{SUPERT}$^{\text{¤}\varepsilon}_2$ & 92.2 \\
\cline{2-8}
& \texttt{CX} & \texttt{DepthScore}$^{\Xi\varepsilon}$ & 95.6 & \texttt{BERTScore}$^{\Xi\varepsilon}_R$ & 95.5 & \texttt{Compression}$^{\text{¤§}}$ & 94.3 \\
\bottomrule
\end{tabular}
\caption{Top 3 metrics per criterion per level (story or system) of absolute Pearson ($r$) correlation. Indices denote different variants.}
\label{tab:top3_metrics}
\end{table*}

\section{Evaluation of automatic metrics using {\NAME}}
\label{sec:analysis_metrics}
In this section we evaluate how suitable existing automatic metrics are for ASG evaluation, using the \texttt{SummEval} library \citep{fabbri2021summeval}\footnote{\url{https://github.com/Yale-LILY/SummEval}}. Due to space constraints, in each figure, we selected representative metrics from each of the categories introduced in \autoref{par:automatic_evaluation}. Full figures can be found in the appendix.

\subsection{Correlations with human judgement}

\begin{figure*}[!ht]
\centering
\begin{minipage}{0.47\linewidth}
    \centering
    \includegraphics[width=\columnwidth]{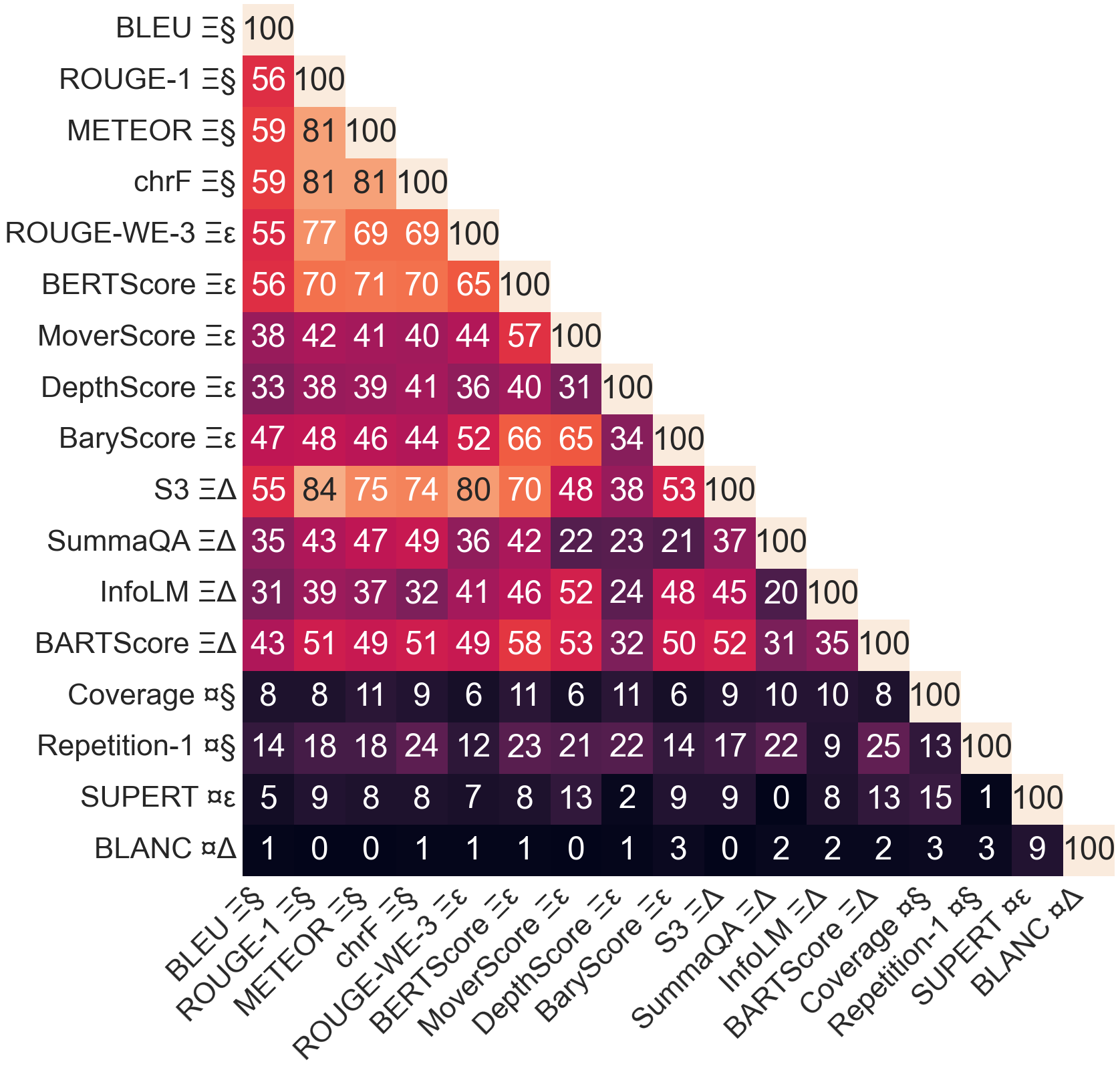}
    \caption{Story-level absolute Kendall correlations (\%) between metrics. Full version: \autoref{fig:story_level_metric_correlations_kendall}.}
    \label{fig:story_level_metrics2_correlations_kendall}
\end{minipage}
\hspace{0.3cm}
\begin{minipage}{0.47\linewidth}
    \centering
    \includegraphics[width=\columnwidth]{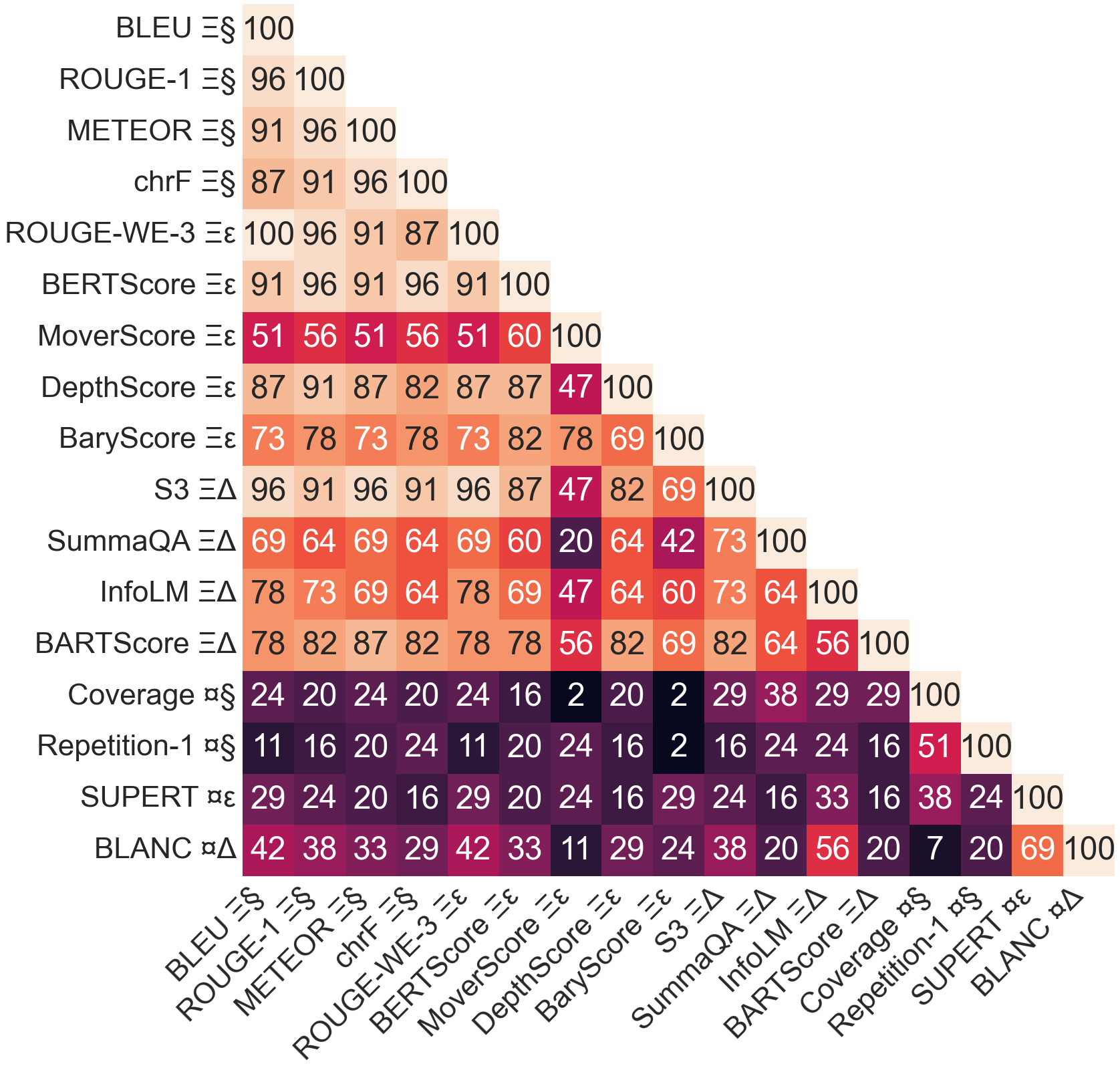}
    \caption{System-level absolute Kendall correlations (\%) between metrics. Full version: \autoref{fig:system_level_metric_correlations_kendall}.}
    \label{fig:system_level_metrics2_correlations_kendall}
\end{minipage}
\end{figure*}

\noindent \textbf{Story-level analysis (\autoref{fig:story_level_mixed2_correlations_kendall})}. Most metrics have either a moderate (between 30\% and 50\%) or weak (below 30\%) correlation with human criteria. \texttt{RE} is particularly elusive, except for the \texttt{SUPERT}$^{\text{¤}\varepsilon}$ metric, which is reference-free and compares the prompt and the story. This corroborates \citet{novikova2017we}, who argue that automatic metrics do not accurately reflect human judgment when comparing instances despite performing reliably at the system level. We also observe vertical ``color stripes'': metric performance is consistent across criteria. A weak metric will correlate poorly with all criteria, whereas a more robust metric will be uniformly better.\\
\noindent \textbf{System-level analysis (\autoref{fig:system_level_mixed2_correlations_kendall})}. Correlations are indeed higher at the system-level, hovering between 40\% and 70\% for most metrics. Therefore, while metrics are poor estimators of human criteria at the story level, they can be used to compare systems with reasonable accuracy.\\
\noindent \textbf{Best metrics per criterion (\autoref{tab:top3_metrics})}. We observe that a few metrics are heavily represented in the top 3 for each level. Pretrained transformer-based metrics achieve strong results. The metrics that correlate best with human criteria at the system level are all reference-based: \texttt{ROUGE-S*}$^{\Xi\text{§}}$, \texttt{BaryScore}$^{\Xi\varepsilon}$, \texttt{DepthScore}$^{\Xi\varepsilon}$, and \texttt{BARTScore}$^{\Xi\Delta}_1$. At the story level, \texttt{BARTScore}$^{\text{¤}\Delta}_2$, \texttt{Novelty-1}$^{\text{¤§}}$ and \texttt{Repetition-3}$^{\text{¤§}}$ are reference-free while \texttt{chrF}$^{\Xi\text{§}}$, \texttt{BERTScore}$^{\Xi\varepsilon}$, \texttt{S3}$^{\Xi\Delta}$ are reference-based. As \texttt{Novelty-1} and \texttt{Repetition-3} are simple data statistics, their outperforming all metrics for \texttt{SU} and \texttt{CH} respectively highlights the shortcomings of current metrics.

\subsection{Correlations between automatic metrics}

\begin{figure}[h]
\centering
\includegraphics[width=\columnwidth]{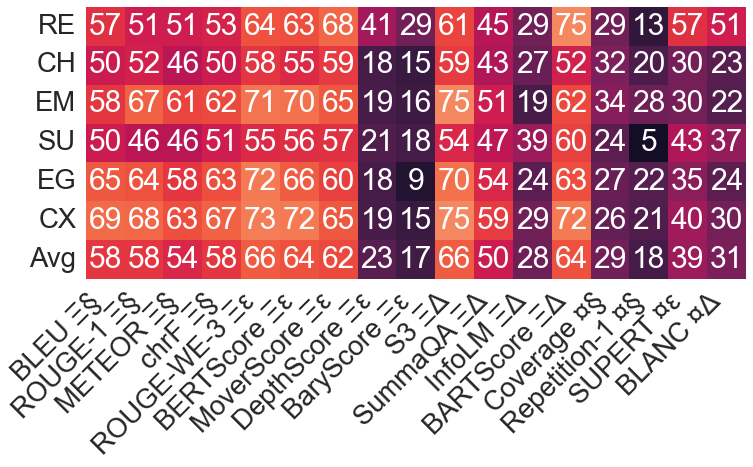}
\caption{Weighted macro F1-scores of paired bootstrap resampling. Full version: \autoref{fig:fscores2}.}
\label{fig:fscores}
\end{figure}

\begin{figure*}[h]
\centering
\includegraphics[width=\textwidth]{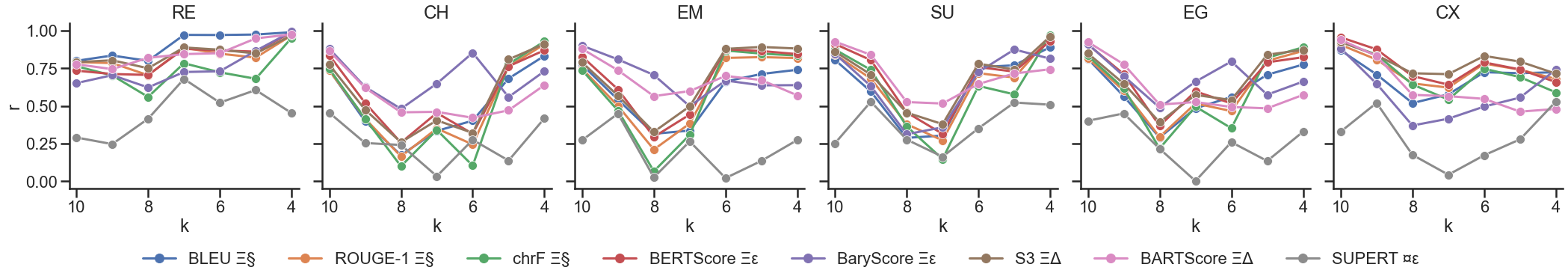}
\caption{System-level absolute Pearson correlations (\%) between automatic metrics and our proposed human criteria on top-$k$ systems}
\label{fig:topk_analysis}
\end{figure*}

\noindent \textbf{Story-level analysis (\autoref{fig:story_level_metrics2_correlations_kendall})}. Reference-based metrics are moderately to highly correlated with one another. By contrast, embedding- and model-based reference-free metrics such as \texttt{SUPERT}$^{\text{¤}\varepsilon}$ and \texttt{BLANC}$^{\text{¤}\Delta}$ are almost independent from all other metrics, even other reference-free metrics.\\
\noindent \textbf{System-level analysis (\autoref{fig:system_level_metrics2_correlations_kendall})}. Previous observations at the story level remain mostly valid, although correlations are overall higher. Reference-based metrics form a large group of very highly correlated metrics, with a majority of correlations surpassing 70\%.
Embedding- and model-based reference-free metrics remain weakly correlated to other metrics.

\subsection{Fine-grained analysis}
\noindent \textbf{Top-$k$ systems (\autoref{fig:topk_analysis})}. Here, we evaluate whether automatic metrics can reliably quantify differences between systems of competitive performances. For all criteria except \texttt{RE} and \texttt{CX}, correlations follow a convex curve between $k=10$ and $k=4$, suggesting that metrics should not be used to compare systems of high variance in quality unless there are enough of them. Indeed, removing a few systems causes correlations to worsen significantly, until the remaining systems are few enough and of competitive performance. \texttt{RE} correlations interestingly increase as $k$ decreases, which indicates that system quantity is a lesser concern for \texttt{RE}.

\noindent \textbf{Pairwise system comparison (\autoref{fig:fscores})}. Here, we evaluate the pairwise discrimative power of automatic metrics. Following \citet{bhandari2020re}, we take all system pairs $(s_1, s_2)$ and compute their average ratings per criterion using paired bootstrap resampling \citep{koehn-2004-statistical, dror-etal-2018-hitchhikers}. We assign a label $y_\mathrm{true} = 1$ if $s_1$ is better than $s_2$ with 95\% confidence, $y_\mathrm{true} = 2$ if $s_2$ is better, and $y_\mathrm{true} = 0$ if confidence is below 95\%. We then repeat the procedure for each metric $m$, getting $y_\mathrm{pred}^{(m)}$ labels, and calculate the weighted macro F1-scores \citep{goutte2005probabilistic} between $y_\mathrm{true}$ and $y_\mathrm{pred}^{(m)}$ to evaluate if $m$ is a good proxy for human criteria. We observe that reference-based metrics again perform better than reference-free metrics, with \texttt{S3}$^{\Xi\Delta}$ and \texttt{ROUGE-WE-3}$^{\Xi\varepsilon}$ at the top. \texttt{DepthScore}$^{\Xi\varepsilon}$ and \texttt{BaryScore}$^{\Xi\varepsilon}$ prove to be very unsuited for pairwise system comparisons, despite showing high system-level correlations (see \autoref{fig:story_level_mixed2_correlations_kendall}). Finally, \texttt{SU} appears to be the most troublesome criterion for this task, suggesting that the surprise factor is especially difficult to evaluate.

\noindent \textbf{Statistical significance.} Using the Williams test (\autoref{sec:williams}), we found that increases in correlation with human criteria between top 3 metrics per criterion (\autoref{tab:top3_metrics}) are not statistically significant, which suggests that best-scoring metrics are of similar performance. However, except for the \texttt{RE} criterion, we notably find that the increases in correlation of \texttt{chrF}$^{\Xi\text{§}}$ and \texttt{BERTScore}$^{\Xi\varepsilon}$ compared to \texttt{BLEU}$^{\Xi\text{§}}$ and \texttt{ROUGE}$^{\Xi\text{§}}$ variants are statistically significant.

\subsection{Aggregated rankings of metrics}
%Here we provide a clear-cut answer to the question of the most appropriate existing metrics.
To aggregate the scores obtained by the three correlation measures (Kendall, Pearson and Spearman), we use the work of \citet{colombo2022best}\footnote{\url{https://github.com/PierreColombo/RankingNLPSystems}}, who rely on the Kemeny consensus \citep{kemeny1959mathematics,myerson1996fundamentals} and recommend to use the Borda Count (\texttt{BC}) as an efficient approximation \citep{sibony2014borda}. They experimentally show that Kemeny consensus has more desirable properties than a ranking obtained through a mean-aggregation procedure. We report  the results in \autoref{tab:top5_metrics_one_level_ranking}. To compare system performance, model- or embedding-based metrics (\textit{e.g..}\ \texttt{BARTScore}$^{\Xi\varepsilon}$ or \texttt{BERTScore}$^{\Xi\varepsilon}$) seem most adapted. However, at the story level, \texttt{chrF}$^{\Xi\text{§}}$ and \texttt{BERTScore}$^{\Xi\varepsilon}$ are among the best metrics, while \texttt{BLEU}$^{\Xi\text{§}}$ is completely absent from the top spots. \texttt{ROUGE}$^{\Xi\text{§}}$ does appear in the ranking, albeit below \texttt{chrF}$^{\Xi\text{§}}$.% We strongly advise to use human annotations.\\
% \noindent \textbf{Takeaways}. Awaiting specific ASG metrics, researchers should use better metrics than \texttt{BLEU}$^{\Xi\text{§}}$ and \texttt{ROUGE}$^{\Xi\text{§}}$. \texttt{chrF}$^{\Xi\text{§}}$ and \texttt{BARTScore}$^{\Xi\varepsilon}$ are the best performers at the story- and system-level respectively. Given the overall weak results, we strongly advise to rely on human annotations for ASG evaluation.

\begin{table}[h]
\centering
\small
\begin{tabular}{clr}
\toprule
Level & Metric & \multicolumn{1}{l}{\texttt{BC}} \\
\midrule
\multirow{5}{*}{Story} & \texttt{chrF}$^{\Xi\text{§}}$ & 1237 \\
& \texttt{S3}$^{\Xi\Delta}_p$ & 1198 \\
& \texttt{ROUGE-1}$^{\Xi\text{§}}$ & 1186 \\
& \texttt{S3}$^{\Xi\Delta}_r$ & 1177 \\
& \texttt{BERTScore}$^{\Xi\varepsilon}_R$ & 1158 \\
\midrule
\multirow{5}{*}{System} & \texttt{BARTScore}$^{\Xi\Delta}$ & 1120\\
& \texttt{BaryScore}$^{\Xi\varepsilon}_5$ & 1110 \\
& \texttt{BERTScore}$^{\Xi\varepsilon}_F$ & 1095 \\
& \texttt{MoverScore}$^{\Xi\varepsilon}$ & 1070 \\
& \texttt{DepthScore}$^{\Xi\varepsilon}$ & 1069 \\
% \midrule
% \multirow{5}{*}{Global} & $\Xi$§\,\texttt{chrF} & 2294 \\
% & $\Xi\Delta$\,\texttt{BARTScore} & 2255 \\
% & $\Xi\Delta$\,\texttt{BERTScore}$_R$ & 2224 \\
% & $\Xi\Delta$\,\texttt{S3}$_p$ & 2174 \\
% & $\Xi\Delta$\,\texttt{S3}$_r$ & 2133 \\
\bottomrule
\end{tabular}
\caption{Top 5 metrics computed by one-level ranking per aggregation level, higher Borda count is better}
\label{tab:top5_metrics_one_level_ranking}
\end{table}

\section{Conclusions}

Our analysis yields the following conclusions:

\begin{enumerate}[wide, labelindent=0pt, noitemsep]
    \item \textbf{Large pre-trained language models seem to produce the best results for ASG.} Our benchmark shows that \texttt{GPT-2} performed better than systems specifically tailored for ASG despite being older than some of them. Overall, all systems remain significantly inferior to human output, illustrating that ASG remains a challenging task for current language models.
    \item \textbf{Stronger metrics, tailored explicitly for specific criteria of ASG, are desperately needed.} The weak correlations of automatic metrics with human criteria still leave much to be desired. Ideally, we would have automatic metrics which reflect each of our proposed criteria.
    \item \textbf{Awaiting specific ASG metrics, researchers should use better metrics than \texttt{BLEU}$^{\Xi\text{§}}$ and \texttt{ROUGE}$^{\Xi\text{§}}$}. \texttt{chrF}$^{\Xi\text{§}}$ and \texttt{BARTScore}$^{\Xi\varepsilon}$ are the best performers at the story- and system-level respectively. Given the overall weak results, however, we strongly advise to rely on human annotations for ASG evaluation.
    \item \textbf{Our new set of human criteria allows for a standardized and extensive human evaluation.} The criteria are overall weakly correlated with one another, which shows that they are non-redundant, and produce coherent system rankings.
\end{enumerate}
\paragraph{Future directions.} Motivated by our search for  %Similarly to how we drew our 
human criteria from the social science literature, we reckon more collaboration between the NLP and social science communities may yield valuable insights into the question of how to computationally capture good indicators of story quality. In this spirit, we hope that {\NAME} will pave the way for further progress in the evaluation of ASG.

\section*{Acknowledgements}
\label{sec:acknowledgements}
We thank Yejin Choi, Richard Bai, Le Fang, Jian Guan, Hannah Rashkin, David Wilmot and Eden Bensaid for answering to our requests for data.\\
This work was granted access to the HPC resources of IDRIS under the allocation 2022-101838 made by GENCI and was partially funded by the grant ANR-20-CHIA-0012-01 (``NoRDF'').

\newpage
\bibliography{acl}
\bibliographystyle{acl_natbib}

\input{appendix_all}

\end{document}

%% file: appendix_all.tex
\appendix
\clearpage
\addcontentsline{toc}{section}{Appendix} % Add the appendix text to the document TOC
\part{Appendix} % Start the appendix part
\parttoc % Insert the appendix TOC

\section{Amazon Mechanical Turk experiment details}
\label{sec:amt_experiment_instructions}
To complement section \ref{sub:amt_experiment}, the details of the instructions we gave in our Amazon Mechanical Turk experiment can be found in \autoref{tab:amt_experiment_instructions} below.

\section{Names of metric variants}
Here we define the names we give to some variants of the automatic metrics we used.\\
\noindent \texttt{SUPERT} and \texttt{BLANC} are summarization metrics which normally require a source document and a summary. In our setting, we have a prompt and a generated story. The suffix \texttt{PS} means we used the ``Prompt as the Summary'', and \texttt{SS} means the ``Story as the Summary''. The \texttt{Golden} suffix means we used the reference human story as the source document and the generated story as the summary.\\
\noindent Given a couple of texts $(x,y)$, \texttt{BARTScore} computes a score based on the log probability of $y$ given $x$. We used the suffixes \texttt{SH} for (Story, Human), \texttt{HS} for (Human, Story), \texttt{SP} for (Story, Prompt) and \texttt{PS} for (Prompt, Story).\\
\noindent All other names are defined in their respective papers.

\onecolumn
\begin{longtable}[h]{p{0.2\linewidth}p{0.75\linewidth}}
\label{tab:amt_experiment_instructions}\\
\toprule
\multicolumn{2}{c}{\Large Amazon Mechanical Turk example task}\\
\midrule
\multicolumn{2}{p{\textwidth}}{Please read the prompt, the human story and the subject story (both stories might be the same). The story you will have to rate is the \textbf{subject story}.}\\
\multicolumn{2}{p{\textwidth}}{\textbf{Important}: we will reject HITs which were done in \textbf{fewer than 30 seconds} (unless both stories are exceptionally short). Please rest assured: if you take the work seriously, we have no reason to reject it.}\\
\multicolumn{2}{p{\textwidth}}{\textbf{Disclaimer}: some stories have been automatically generated and might contain explicit or offensive content.}\\
\multicolumn{2}{p{\textwidth}}{\textbf{Note}: some stories have been abruptly cut in the middle of a sentence. Please rate them as if they ended just before the unfinished sentence.}\\
\multicolumn{2}{p{\textwidth}}{\textbf{Note}: if the story is not relevant with respect to the prompt, it \textbf{only} affects the \textbf{Relevance} criterion! Do not rate 1 everywhere, or we will \textbf{reject}!}\\
\multicolumn{2}{p{\textwidth}}{Then, please write down the name of the \textbf{first character} that is mentioned in the \textbf{subject story}; if no name is mentioned, write “None”. Only proper nouns count as names.}\\
\multicolumn{2}{p{\textwidth}}{Then, please rate the \textbf{subject story} on a scale from 1 (worst) to 5 (best) on the following criteria: relevance, coherence, empathy, surprise, engagement, and complexity.}\\
\midrule
Prompt & When you die the afterlife is an arena where you face every insect and animal you killed in your life. If you win you go to heaven, lose you go to hell. Your job was an exterminator on earth.\\
\midrule
Human story & 3,000 years have I been fighting. Every morning, the raccoons scratch at my eyes. Every evening, the skunks spray me while the opossums chew at my feet. I have never had any tools. I have only my hands. I don’t remember the place I came from before this. All I remember is the daily fight between me and these animals. No matter how many times I kill them, they come back the next day. [...]\\
\midrule
Subject story & First of all, not everyone was entitled to be an exterminator. But the ones that were – maybe were, like, *genius*, because, yes, I had once belonged to a less fortunate class of people – had all the opportunity to work for the damn plant killer, and it's hard work. And the horrifying truth is, once you die, and the entire planet turns into a glade that contains a golden fish that would've been crushed by a million million goldfish just moments ago, you're not really good enough for heaven. Why? [...]\\
\midrule
Name of the first mentioned character in the subject story & \texttt{[Area to fill]}\\
\midrule
\multirow{5}{*}{\parbox{\linewidth}{Relevance — measures how well the story matches its prompt}} & 1 — The story has no relationship with the prompt at all. \\
& 2 — The story only has a weak relationship with the prompt.\\
& 3 — The story roughly matches the prompt.\\
& 4 — The story matches the prompt, except for one or two small aspects.\\
& 5 — The story matches the prompt exactly.\\
\midrule
\pagebreak
\midrule
\multirow{5}{*}{\parbox{\linewidth}{Coherence — measures whether the story makes sense}} & 1 — The story does not make sense at all. For instance, the setting and/or characters keep changing, and/or there is no understandable plot.\\
& 2 — Most of the story does not make sense.\\
& 3 — The story mostly makes sense but has some incoherences.\\
& 4 — The story almost makes sense overall, except for one or two small incoherences.\\
& 5 — The story makes sense from beginning to end.\\
\midrule
\multirow{5}{*}{\parbox{\linewidth}{Empathy — measures how well you understood the characters' emotions (regardless of whether you agreed with them)}} & 1 — The characters seemed apathetic to you.\\
& 2 — At least one character slightly related to you on an emotional level.\\
& 3 — You recognized specific, but not necessarily strong, emotions (eg sadness, joy, fear…) in at least one character.\\
& 4 — At least one character emotionally involved you, but minor details prevented you from completely relating to them.\\
& 5 — At least one character completely involved you on an emotional level.\\
\midrule
\multirow{5}{*}{\parbox{\linewidth}{Surprise — measures how surprising the end of the story was}} & 1 — The ending seemed completely obvious from the start, or doesn't make any sense at all.\\
& 2 — The ending was easily predictable after a few sentences.\\
& 3 — The ending was predictable after half of the story.\\
& 4 — The ending surprised you, but would have been difficult to predict.\\
& 5 — The ending surprised you, and still seemed as if it could very reasonably have been predicted, ie, there were enough clues in the story.\\
\midrule
\multirow{5}{*}{\parbox{\linewidth}{Engagement — measures how much you engaged with the story}} & 1 — You found the story boring and were glad it was over.\\
& 2 — You found one or two things interesting in the story, but no more.\\
& 3 — The story was mildly interesting.\\
& 4 — The story almost kept you engaged until the end.\\
& 5 — You were so engaged that you wished there was a sequel.\\
\midrule
\multirow{5}{*}{\parbox{\linewidth}{Complexity — measures how elaborate the story is}} & 1 — The setting of the story is extremely simple; it only involves one or two characters or concepts. \\
& 2 — The setting of the story is simple; one or two characters, a simple plot, maybe an indication of time or location.\\
& 3 — The story is somewhat developed: it involves at least one of the following: complex concepts, realistic characters, an intricate plot, an underlying history or circumstances, precise descriptions.\\
& 4 — The story is developed: it involves at least two of the following: complex concepts, realistic characters, an intricate plot, an underlying history or circumstances, precise descriptions.\\
& 5 — The story is well thought-out: it involves at least three of the following: complex concepts, realistic characters, an intricate plot, an underlying history or circumstances, precise descriptions.\\
\bottomrule
\caption{Example task from our Amazon Mechanical Turk experiment}
\end{longtable}

\clearpage
\begin{minipage}{0.45\linewidth}
\section{Distributions of human annotations per system}
Here we report the violin plots of the distributions of human annotations per system. Human output scores visibly better than language models. Note that for our generation, we do not use beam search \citep{colombo2021beam,colombo2020guiding,pichler2022differential,colombo2021learning,colombo2022learning,garcia2019token, colombo2021novel}. To further improve the generation a domain pre-trained language model could be considered \citep{chapuis-etal-2020-hierarchical,colombo2021code}.
\end{minipage}

\begin{figure*}[h!]
    \centering
    \includegraphics[width=\textwidth]{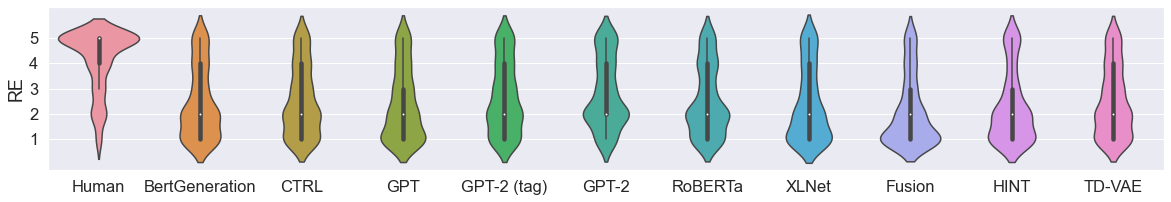}
    \includegraphics[width=\textwidth]{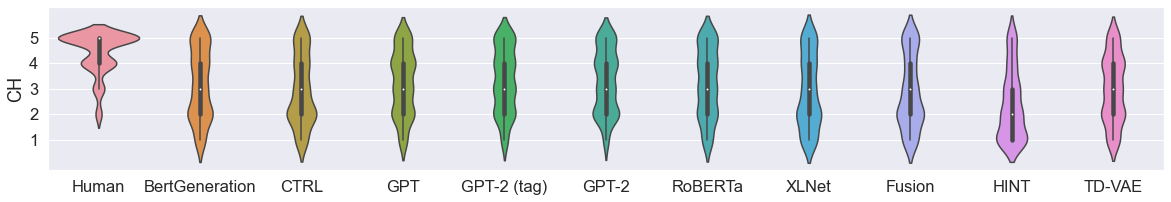}
    \includegraphics[width=\textwidth]{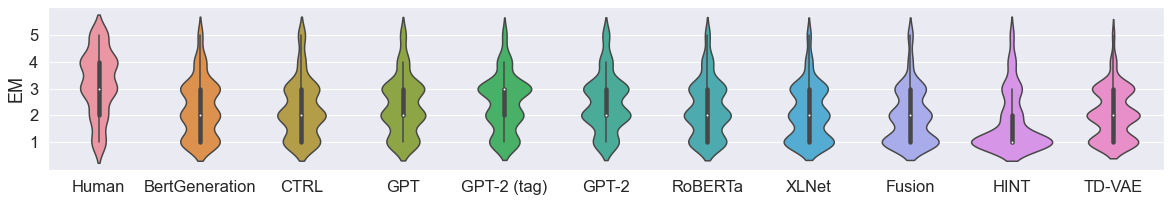}
    \includegraphics[width=\textwidth]{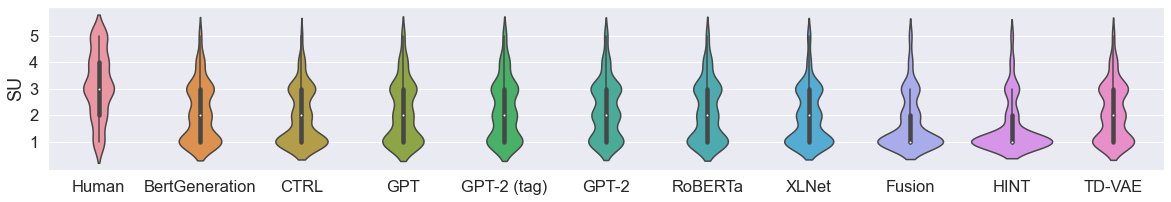}
    \includegraphics[width=\textwidth]{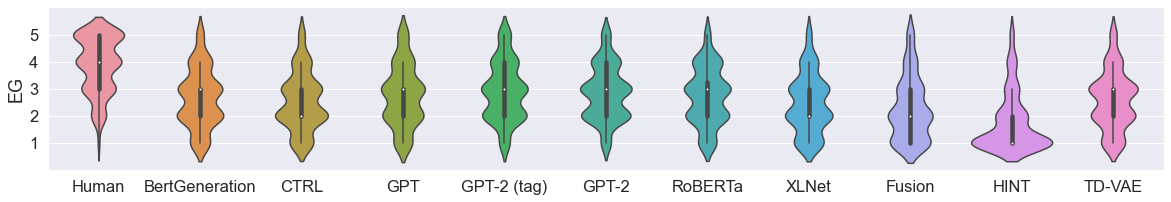}
    \includegraphics[width=\textwidth]{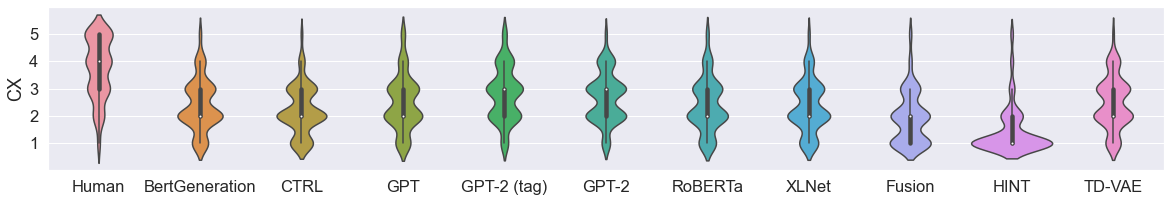}
    \caption{Violin plots of the distributions of human annotations per system}
    \label{fig:violin}
\end{figure*}

\begin{figure}[h]
\centering
\begin{minipage}{0.45\linewidth}
    \centering
    \includegraphics[width=\textwidth]{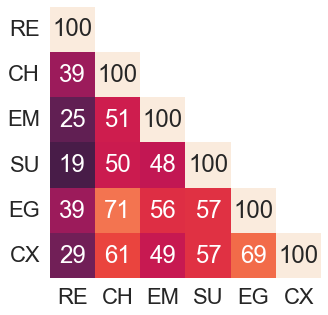}
    \caption{Story-level Spearman correlations (\%) between our proposed human criteria}
    \label{fig:story_level_human_correlations_spearman}
\end{minipage}
\hspace{0.5cm}
\begin{minipage}{0.45\linewidth}
    \centering
    \includegraphics[width=\textwidth]{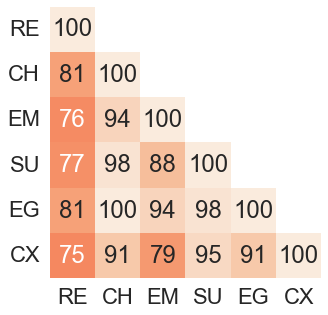}
    \caption{System-level Spearman correlations (\%) between our proposed human criteria}
    \label{fig:system_level_human_correlations_spearman}
\end{minipage}
\end{figure}

\begin{figure}[h]
\centering
\begin{minipage}{0.45\linewidth}
    \centering
    \includegraphics[width=\textwidth]{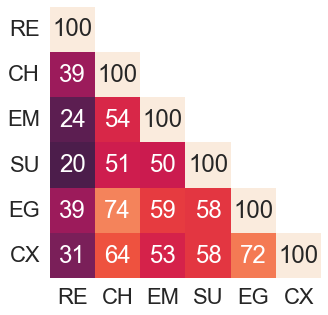}
    \caption{Story-level Pearson correlations (\%) between our proposed human criteria}
    \label{fig:story_level_human_correlations_pearson}
\end{minipage}
\hspace{0.5cm}
\begin{minipage}{0.45\linewidth}
    \centering
    \includegraphics[width=\textwidth]{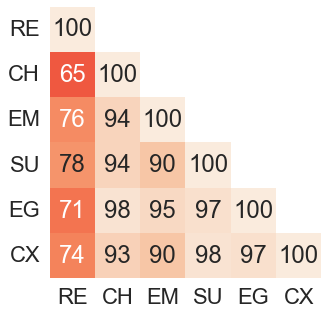}
    \caption{System-level Pearson correlations (\%) between our proposed human criteria}
    \label{fig:system_level_human_correlations_pearson}
\end{minipage}
\end{figure}

\clearpage
\begin{minipage}{0.45\linewidth}
\section{Correlations between human criteria}
Here we report the story-level and system-level absolute correlations between human criteria with Spearman's $\rho$ (\autoref{fig:story_level_human_correlations_spearman} and \autoref{fig:system_level_human_correlations_spearman}) and Pearson's $r$ (\autoref{fig:story_level_human_correlations_pearson} and \autoref{fig:system_level_human_correlations_pearson}).
\end{minipage}

\clearpage
\begin{minipage}{0.45\linewidth}
\section{Correlations between human criteria and automatic metrics}
Here we report the full figures of story-level and system-level absolute correlations between human criteria and automatic metrics with all three correlation coefficients.
\end{minipage}

\begin{figure}[h]
    \centering
    \includegraphics[width=\textwidth]{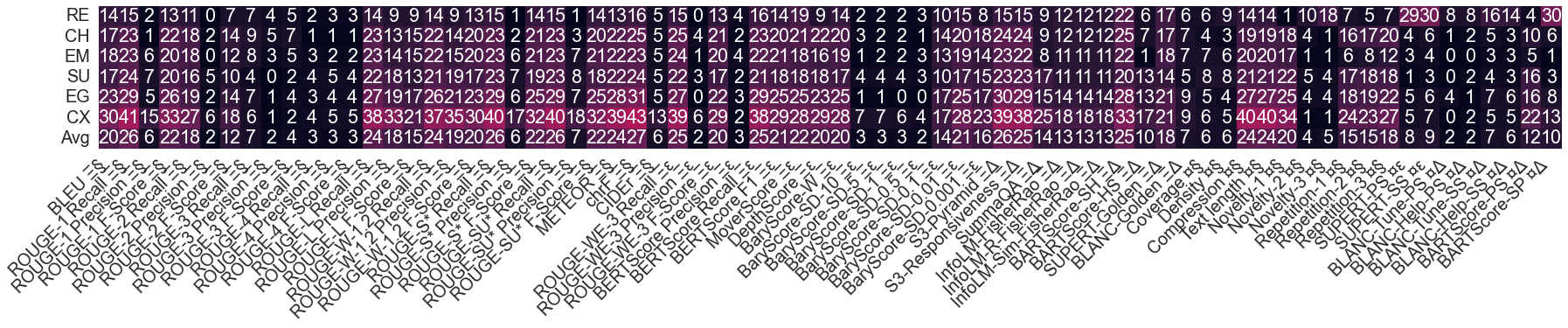}
    \caption{Story-level absolute Kendall correlations (\%) between automatic metrics and our proposed human criteria}
    \label{fig:story_level_mixed_correlations_kendall}
\end{figure}

\begin{figure}[h!]
    \centering
    \includegraphics[width=\textwidth]{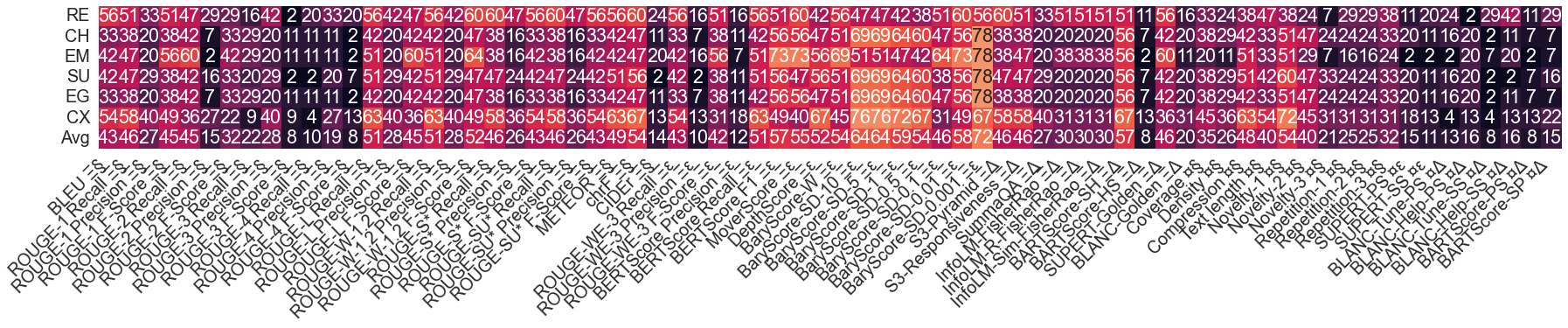}
    \caption{System-level absolute Kendall correlations (\%) between automatic metrics and our proposed human criteria}
    \label{fig:system_level_mixed_correlations_kendall}
\end{figure}

\begin{figure}[h]
    \centering
    \includegraphics[width=\textwidth]{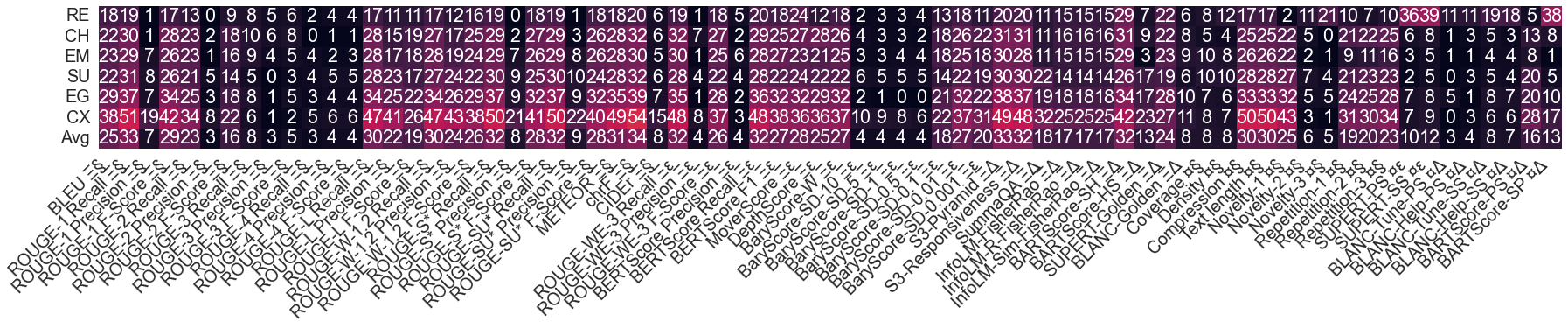}
    \caption{Story-level absolute Spearman correlations (\%) between automatic metrics and our proposed human criteria}
    \label{fig:story_level_mixed_correlations_spearman}
\end{figure}

\begin{figure}[h!]
    \centering
    \includegraphics[width=\textwidth]{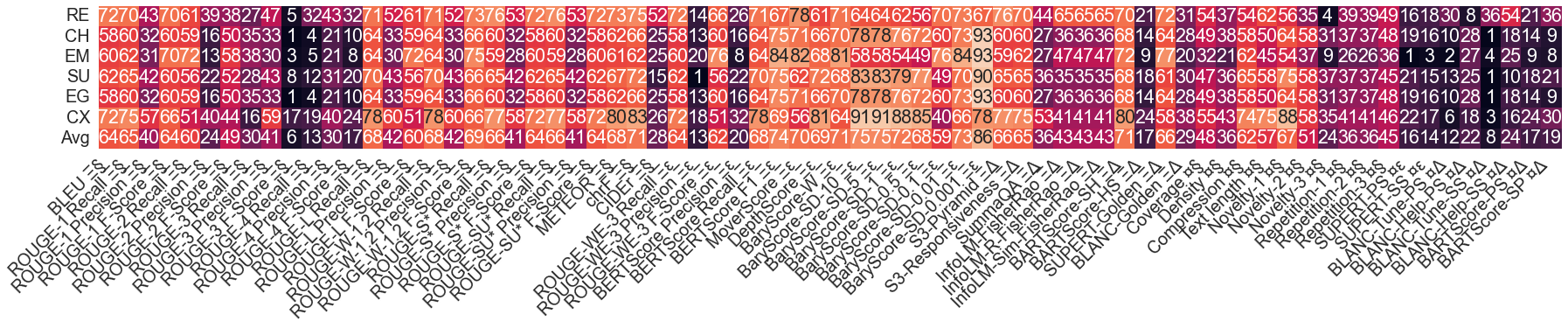}
    \caption{System-level absolute Spearman correlations (\%) between automatic metrics and our proposed human criteria}
    \label{fig:system_level_mixed_correlations_spearman}
\end{figure}

\clearpage
\begin{figure}[h]
    \centering
    \includegraphics[width=\textwidth]{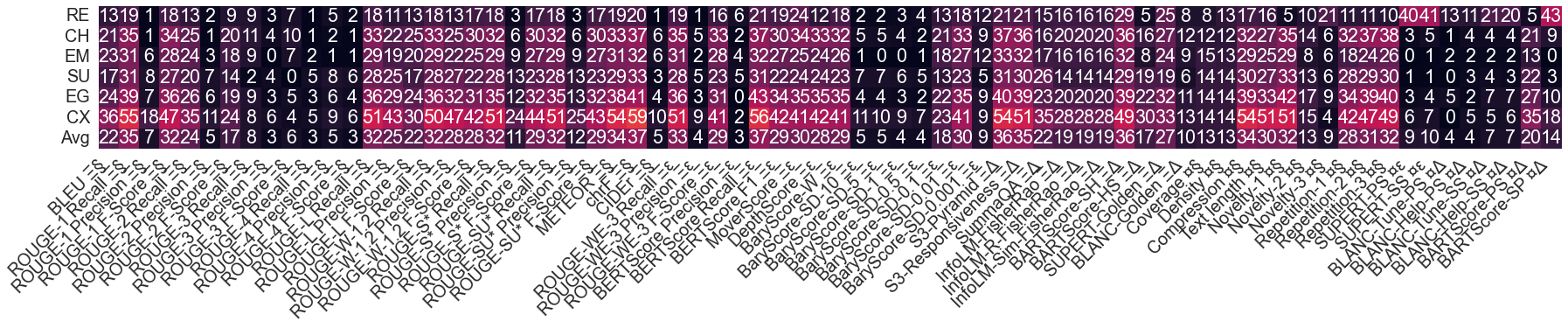}
    \caption{Story-level absolute Pearson correlations (\%) between automatic metrics and our proposed human criteria}
    \label{fig:story_level_mixed_correlations_pearson}
\end{figure}

\begin{figure}[h]
    \centering
    \includegraphics[width=\textwidth]{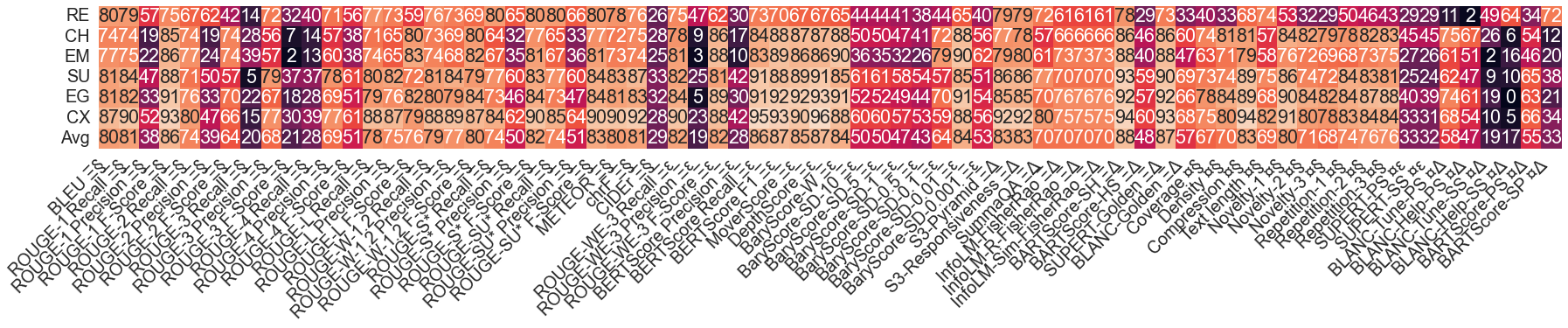}
    \caption{System-level absolute Pearson correlations (\%) between automatic metrics and our proposed human criteria}
    \label{fig:system_level_mixed_correlations_pearson}
\end{figure}

\clearpage
\begin{minipage}{0.45\linewidth}
\section{Correlations between automatic metrics}
Here we report the full figures of story-level and system-level absolute correlations between automatic metrics with all three correlation coefficients.
\end{minipage}

\begin{figure}[h]
    \centering
    \includegraphics[width=\textwidth]{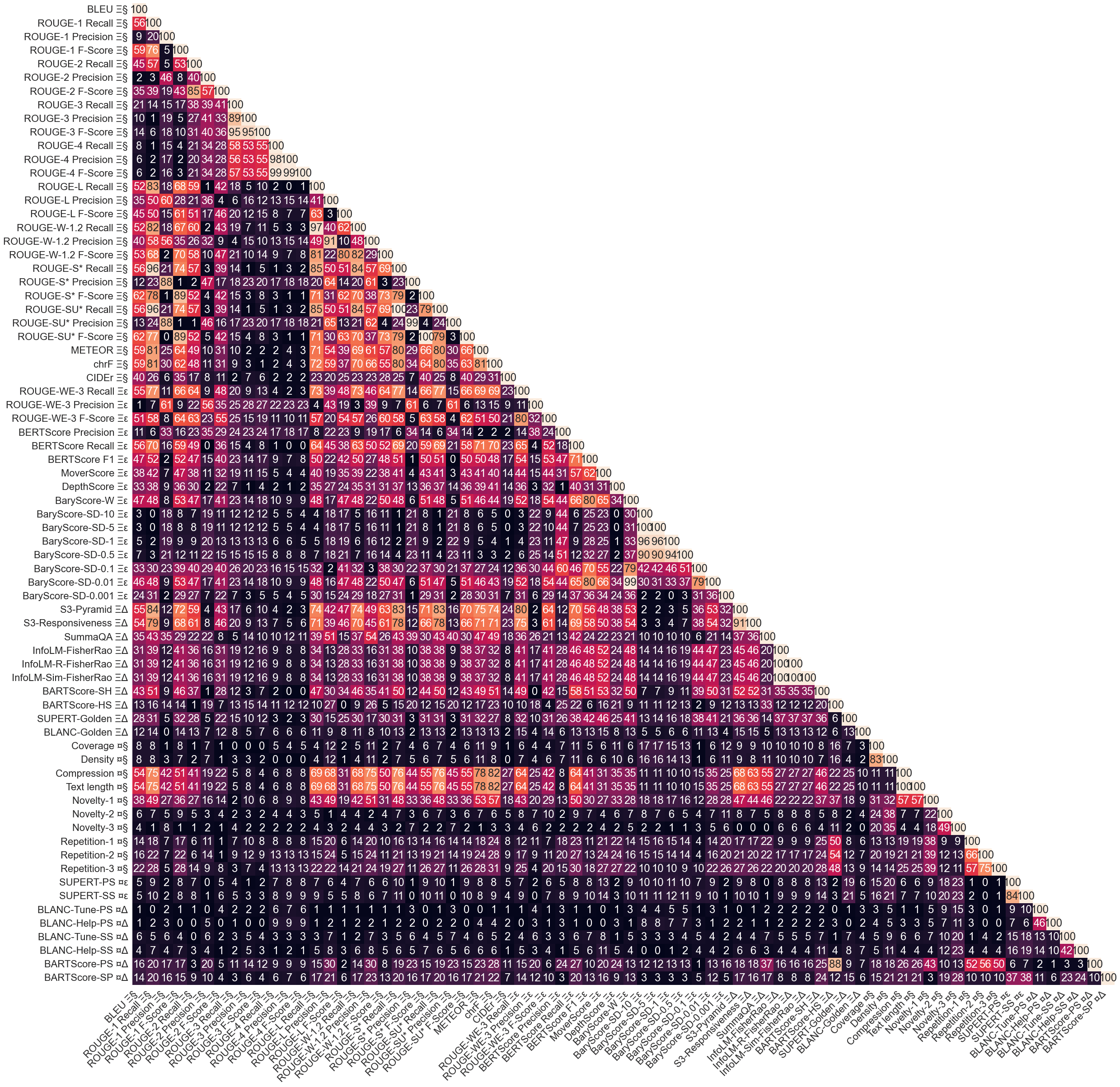}
    \caption{Story-level absolute Kendall correlations (\%) between automatic metrics}
    \label{fig:story_level_metric_correlations_kendall}
\end{figure}

\begin{figure}[h]
    \centering
    \includegraphics[width=\textwidth]{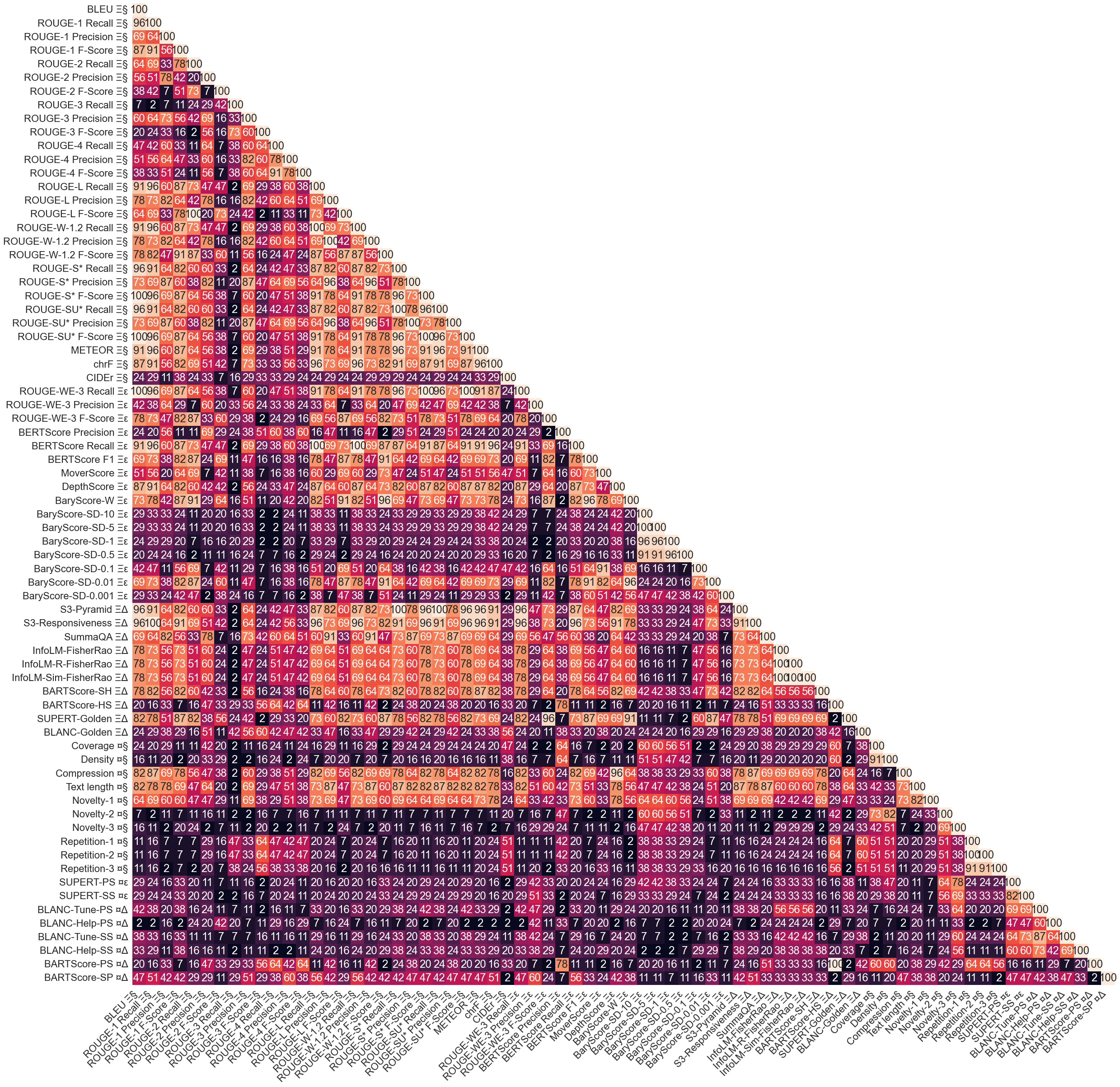}
    \caption{System-level absolute Kendall correlations (\%) between automatic metrics}
    \label{fig:system_level_metric_correlations_kendall}
\end{figure}

\begin{figure}[h]
    \centering
    \includegraphics[width=\textwidth]{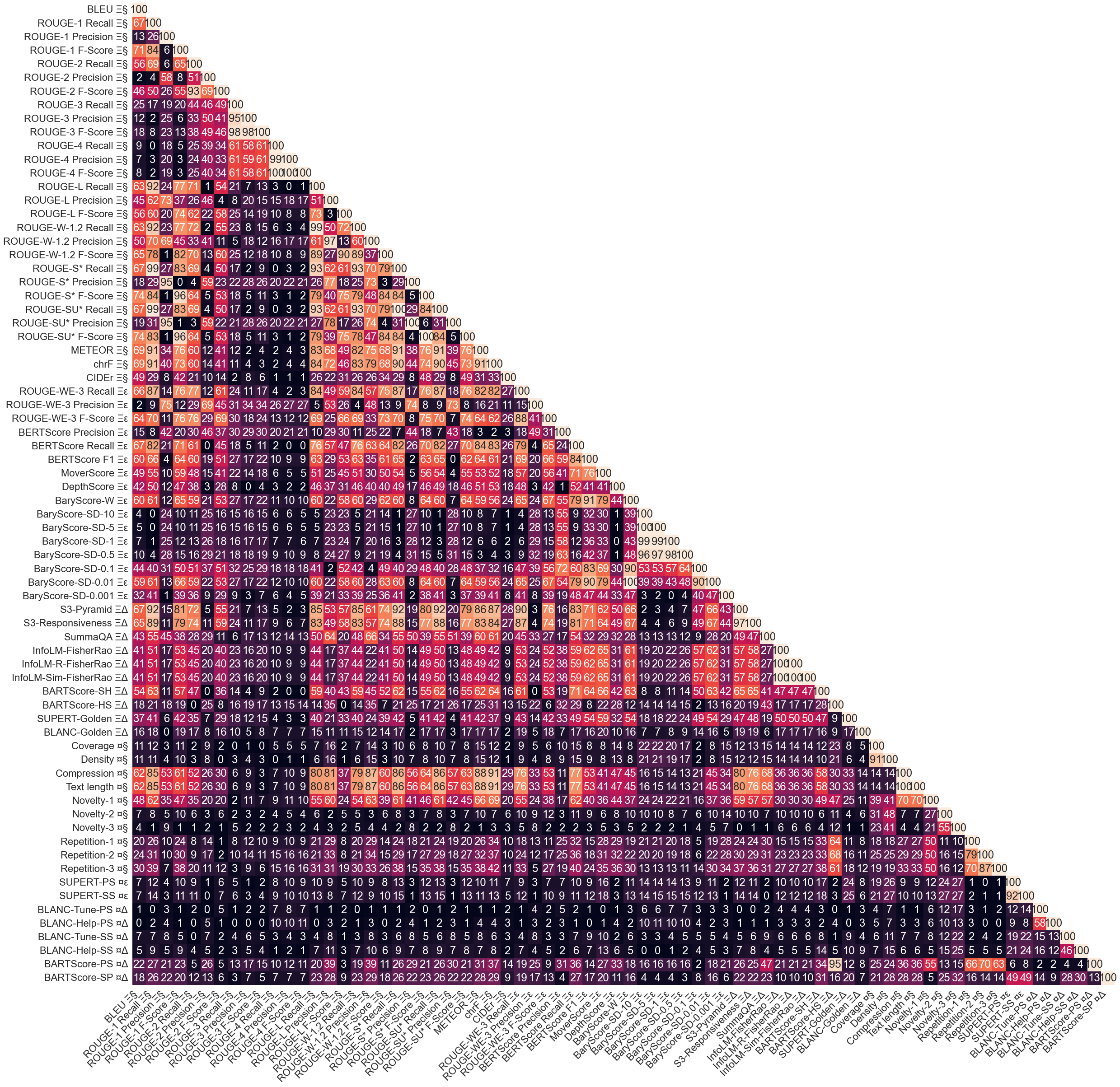}
    \caption{Story-level absolute Spearman correlations (\%) between automatic metrics}
    \label{fig:story_level_metric_correlations_spearman}
\end{figure}

\begin{figure}[h]
    \centering
    \includegraphics[width=\textwidth]{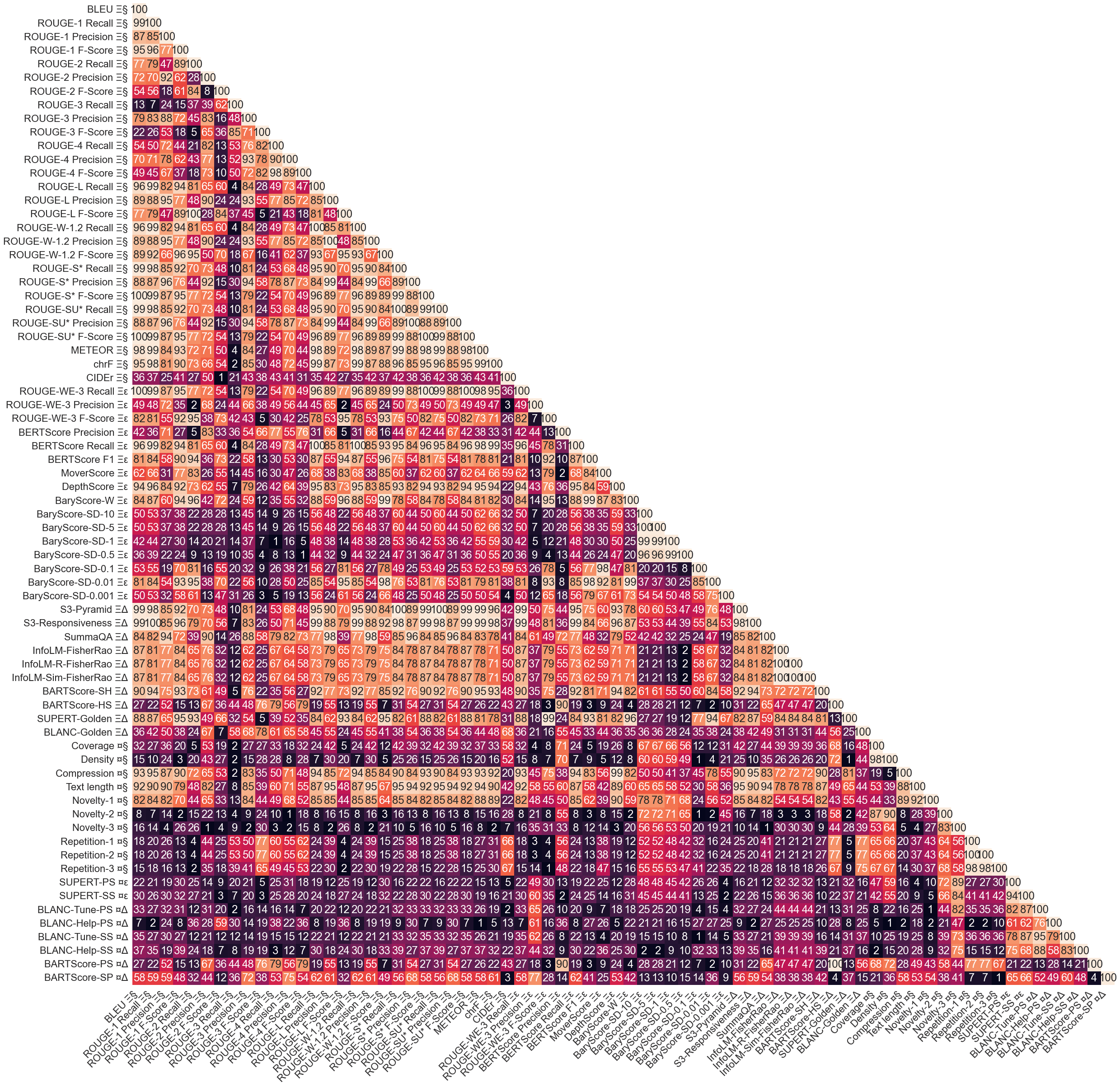}
    \caption{System-level absolute Spearman correlations (\%) between automatic metrics}
    \label{fig:system_level_metric_correlations_spearman}
\end{figure}

\begin{figure}[h!]
    \centering
    \includegraphics[width=\textwidth]{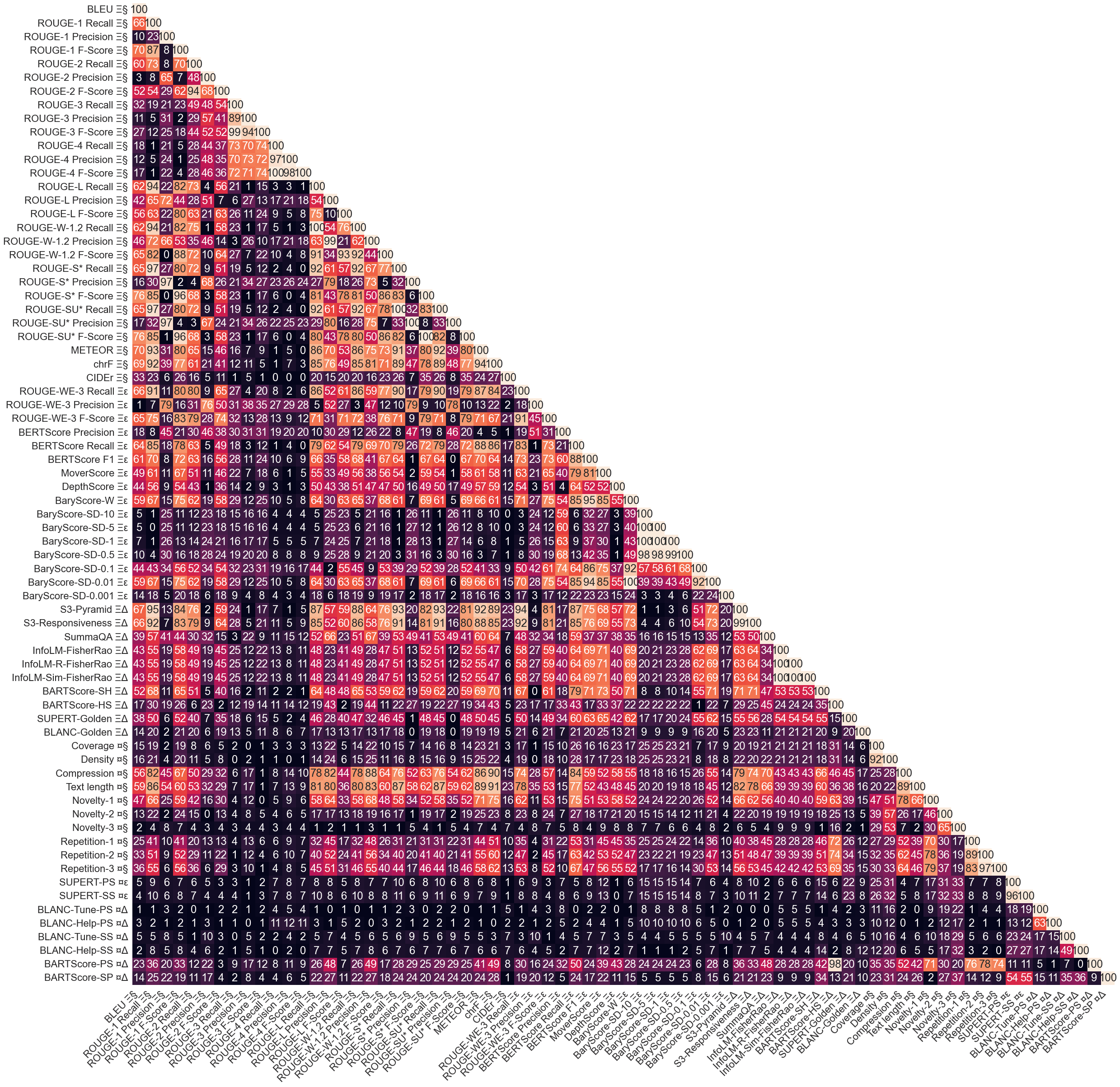}
    \caption{Story-level absolute Pearson correlations (\%) between automatic metrics}
    \label{fig:story_level_metric_correlations_pearson}
\end{figure}

\begin{figure}[h]
    \centering
    \includegraphics[width=\textwidth]{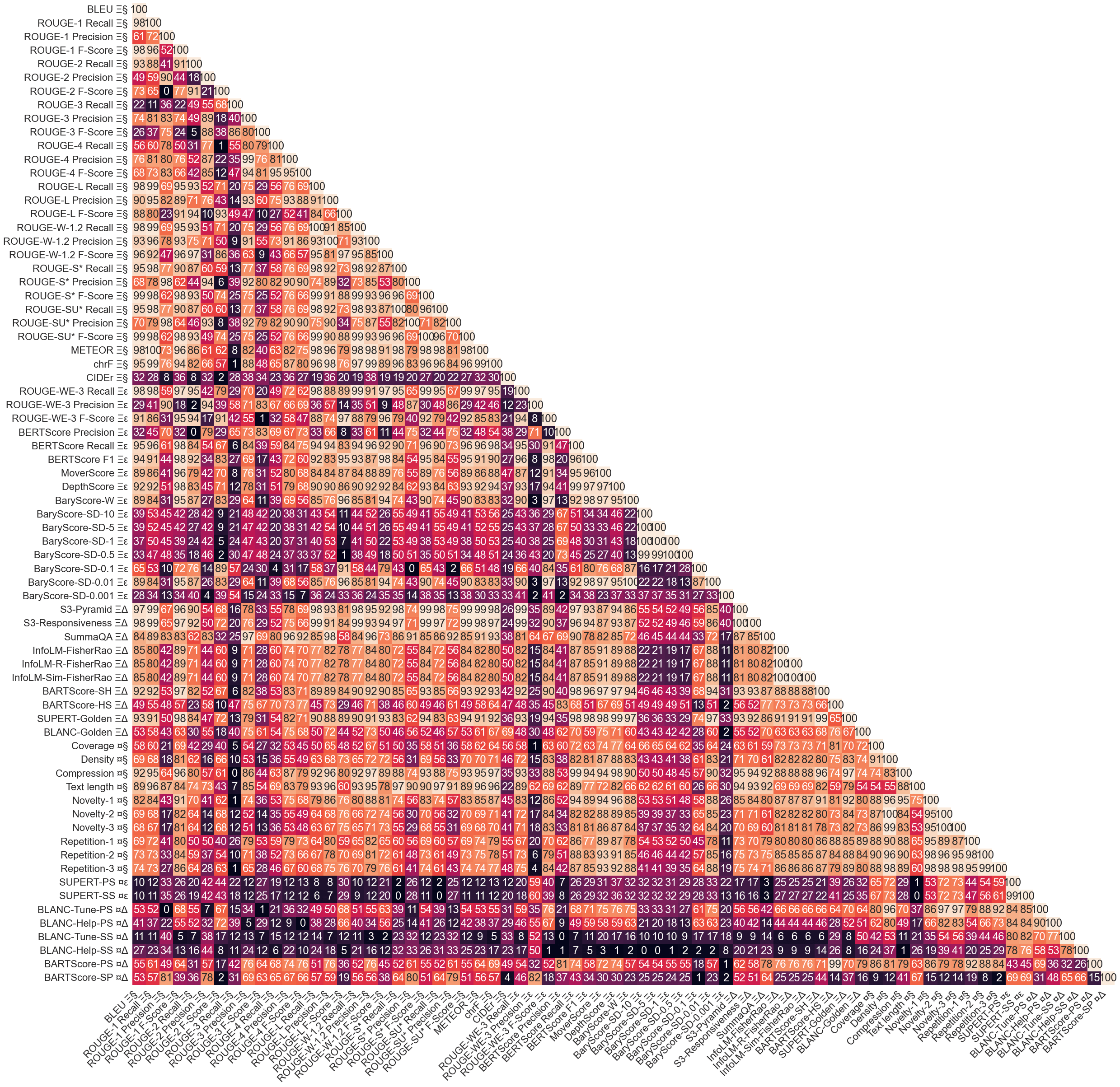}
    \caption{System-level absolute Pearson correlations (\%) between automatic metrics}
    \label{fig:system_level_metric_correlations_pearson}
\end{figure}

\clearpage
\begin{minipage}{0.45\linewidth}
\section{Best metrics per criterion per level of correlation per correlation coefficient}
Here we report the top 5 metrics per criterion per story-level and system-level absolute correlation coefficient.
\end{minipage}

\begin{table}[h!]
\centering \small
\begin{tabular}{lrrrrrr}
\toprule
Criterion & \multicolumn{2}{c}{$|\tau|$ (\%)} & \multicolumn{2}{c}{$|\rho|$ (\%)} & \multicolumn{2}{c}{$|r|$ (\%)} \\
\midrule
\multirow{5}{*}{\texttt{RE}} & SUPERT-SS$^{\Xi\varepsilon}$ & 29.95 & SUPERT-SS$^{\Xi\varepsilon}$ & 38.58 & BARTScore-SP$^{\text{¤}\Delta}$ & 42.55 \\
& BARTScore-SP$^{\text{¤}\Delta}$ & 29.61 & BARTScore-SP$^{\text{¤}\Delta}$ & 37.98 & SUPERT-SS$^{\Xi\varepsilon}$ & 41.16 \\
& SUPERT-PS$^{\Xi\varepsilon}$ & 28.59 & SUPERT-PS$^{\Xi\varepsilon}$ & 36.40 & SUPERT-PS$^{\Xi\varepsilon}$ & 40.15 \\
& BARTScore-SH$^{\Xi\Delta}$ & 22.32 & BARTScore-SH$^{\Xi\Delta}$ & 28.53 & BARTScore-SH$^{\Xi\Delta}$ & 28.98 \\
& MoverScore$^{\Xi\varepsilon}$ & 19.12 & MoverScore$^{\Xi\varepsilon}$ & 23.67 & SUPERT-Golden$^{\Xi\varepsilon}$ & 24.72 \\
\midrule
\multirow{5}{*}{\texttt{CH}} & ROUGE-WE-3 Recall$^{\Xi\varepsilon}$ & 25.29 & ROUGE-WE-3 Recall$^{\Xi\varepsilon}$ & 32.22 & Repetition-3$^{\text{¤§}}$ & 38.12 \\
& BARTScore-SH$^{\Xi\Delta}$ & 25.06 & \textsc{chrF}$^{\Xi\text{§}}$ & 32.03 & BERTScore Recall$^{\Xi\varepsilon}$ & 37.12 \\
& \textsc{chrF}$^{\Xi\text{§}}$ & 24.61 & BARTScore-SH$^{\Xi\Delta}$ & 31.38 & S3-Pyramid$^{\Xi\Delta}$ & 37.05 \\
& S3-Pyramid$^{\Xi\Delta}$ & 24.39 & S3-Responsiveness$^{\Xi\Delta}$ & 31.31 & \textsc{chrF}$^{\Xi\text{§}}$ & 36.99 \\
& S3-Responsiveness$^{\Xi\Delta}$ & 24.28 & S3-Pyramid$^{\Xi\Delta}$ & 31.14 & Repetition-2$^{\text{¤§}}$ & 36.54 \\
\midrule
\multirow{5}{*}{\texttt{EM}} & ROUGE-WE-3 Recall$^{\Xi\varepsilon}$ & 23.58 & ROUGE-WE-3 Recall$^{\Xi\varepsilon}$ & 29.85 & S3-Pyramid$^{\Xi\Delta}$ & 32.78 \\
& \textsc{chrF}$^{\Xi\text{§}}$ & 23.33 & \textsc{chrF}$^{\Xi\text{§}}$ & 29.81 & \textsc{chrF}$^{\Xi\text{§}}$ & 32.43 \\
& S3-Pyramid$^{\Xi\Delta}$ & 23.19& S3-Pyramid$^{\Xi\Delta}$ & 29.68 & BERTScore Recall$^{\Xi\varepsilon}$ & 32.06 \\
& ROUGE-SU* Recall$^{\Xi\text{§}}$ & 23.13 & ROUGE-SU* Recall$^{\Xi\text{§}}$ & 29.38 & S3-Responsiveness$^{\Xi\Delta}$ & 31.78 \\
& ROUGE-S* Recall$^{\Xi\text{§}}$ & 23.08 & ROUGE-S* Recall$^{\Xi\text{§}}$ & 29.32 & BARTScore-SH$^{\Xi\Delta}$ & 31.66 \\
\midrule
\multirow{5}{*}{\texttt{SU}} & \textsc{chrF}$^{\Xi\text{§}}$ & 24.45 & \textsc{chrF}$^{\Xi\text{§}}$ & 31.55 & Novelty-1$^{\text{¤§}}$ & 32.86 \\
& ROUGE-1 Recall$^{\Xi\text{§}}$ & 23.67 & ROUGE-1 Recall$^{\Xi\text{§}}$ & 30.86 & \textsc{chrF}$^{\Xi\text{§}}$ & 32.65 \\
& S3-Responsiveness$^{\Xi\Delta}$ & 23.35 & S3-Responsiveness$^{\Xi\Delta}$ & 30.41 & ROUGE-1 Recall$^{\Xi\text{§}}$ & 31.32 \\
& Novelty-1$^{\text{¤§}}$ & 23.11 & ROUGE-SU* Recall$^{\Xi\text{§}}$ & 30.30 & S3-Pyramid$^{\Xi\Delta}$ & 31.07 \\
& ROUGE-SU* Recall$^{\Xi\text{§}}$ & 22.85 & ROUGE-S* Recall$^{\Xi\text{§}}$ & 30.25 & BERTScore Recall$^{\Xi\varepsilon}$ & 30.98 \\
\midrule
\multirow{5}{*}{\texttt{EG}} & \textsc{chrF}$^{\Xi\text{§}}$ & 30.77 & \textsc{chrF}$^{\Xi\text{§}}$ & 39.03 & BERTScore Recall$^{\Xi\varepsilon}$ & 42.95 \\
& S3-Pyramid$^{\Xi\Delta}$ & 29.62 & S3-Pyramid$^{\Xi\Delta}$ & 37.74 & Novelty-1$^{\text{¤§}}$ & 42.27 \\
& ROUGE-1 Recall$^{\Xi\text{§}}$ & 29.19 & ROUGE-1 Recall$^{\Xi\text{§}}$ & 37.02 & \textsc{chrF}$^{\Xi\text{§}}$ & 41.07 \\
& S3-Responsiveness$^{\Xi\Delta}$ & 29.01 & S3-Responsiveness$^{\Xi\Delta}$ & 36.85 & S3-Pyramid$^{\Xi\Delta}$ & 40.34 \\
& BERTScore Recall$^{\Xi\varepsilon}$ & 28.93 & ROUGE-S* Recall$^{\Xi\text{§}}$ & 36.60 & Repetition-3$^{\text{¤§}}$ & 39.53 \\
\midrule
\multirow{5}{*}{\texttt{CX}} & \textsc{chrF}$^{\Xi\text{§}}$ & 43.31 & \textsc{chrF}$^{\Xi\text{§}}$ & 54.11 & \textsc{chrF}$^{\Xi\text{§}}$ & 58.76 \\
& ROUGE-1 Recall$^{\Xi\text{§}}$ & 40.65 & ROUGE-1 Recall$^{\Xi\text{§}}$ & 50.60 & BERTScore Recall$^{\Xi\varepsilon}$ & 55.83 \\
& ROUGE-SU* Recall$^{\Xi\text{§}}$ & 39.83 & Text length$^{\text{¤§}}$ & 50.19 & ROUGE-1 Recall$^{\Xi\text{§}}$ & 55.01 \\
& Text length$^{\text{¤§}}$ & 39.82 & Compression$^{\text{¤§}}$ & 50.19 & METEOR$^{\Xi\text{§}}$ & 54.41 \\
& Compression$^{\text{¤§}}$ & 39.82 & ROUGE-SU* Recall$^{\Xi\text{§}}$ & 50.10 & Compression$^{\text{¤§}}$ & 54.38 \\
\bottomrule
\end{tabular}
\caption{Top 5 metrics per criterion per story-level correlation coefficient}
\label{tab:top5_story}
\end{table}

\clearpage
\begin{table}[ht!]
\centering \small
\begin{tabular}{lrrrrrr}
\toprule
Criterion & \multicolumn{2}{c}{$|\tau|$ (\%)} & \multicolumn{2}{c}{$|\rho|$ (\%)} & \multicolumn{2}{c}{$|r|$ (\%)} \\
\midrule
\multirow{5}{*}{\texttt{RE}} & S3-Pyramid$^{\Xi\Delta}$ & 60.00 & MoverScore$^{\Xi\varepsilon}$ & 78.18 & ROUGE-S* F-Score$^{\Xi\text{§}}$ & 80.39 \\
& \textsc{chrF}$^{\Xi\text{§}}$ & 60.00 & S3-Pyramid$^{\Xi\Delta}$ & 75.76 & ROUGE-SU* F-Score$^{\Xi\text{§}}$ & 80.29 \\
& ROUGE-SU* Recall$^{\Xi\text{§}}$ & 60.00 & ROUGE-S* Recall$^{\Xi\text{§}}$ & 75.76 & ROUGE-S* Recall$^{\Xi\text{§}}$ & 80.24 \\
& ROUGE-S* Recall$^{\Xi\text{§}}$ & 60.00 & ROUGE-SU* Recall$^{\Xi\text{§}}$ & 75.76 & ROUGE-SU* Recall$^{\Xi\text{§}}$ & 80.23 \\
& ROUGE-W-1.2 F-Score$^{\Xi\text{§}}$ & 60.00 & \textsc{chrF}$^{\Xi\text{§}}$ & 74.55 & BLEU$^{\Xi\text{§}}$ & 79.89 \\
\midrule
\multirow{5}{*}{\texttt{CH}} & BaryScore-SD-0.001$^{\Xi\varepsilon}$ & 77.78 & BaryScore-SD-0.001$^{\Xi\varepsilon}$ & 92.73 & BaryScore-SD-0.01$^{\Xi\varepsilon}$ & 88.15 \\
& BaryScore-SD-5$^{\Xi\varepsilon}$ & 68.89 & BaryScore-SD-5$^{\Xi\varepsilon}$ & 78.18 & BaryScore-W$^{\Xi\varepsilon}$ & 87.99 \\
& BaryScore-SD-10$^{\Xi\varepsilon}$ & 68.89 & BaryScore-SD-10$^{\Xi\varepsilon}$ & 78.18 & BERTScore F1$^{\Xi\varepsilon}$ & 87.91 \\
& BaryScore-SD-1$^{\Xi\varepsilon}$ & 64.44 & BaryScore-SD-1$^{\Xi\varepsilon}$ & 75.76 & DepthScore$^{\Xi\varepsilon}$ & 87.38 \\
& BaryScore-SD-0.5$^{\Xi\varepsilon}$ & 60.00 & BERTScore F1$^{\Xi\varepsilon}$ & 74.55 & MoverScore$^{\Xi\varepsilon}$ & 86.95 \\
\midrule
\multirow{5}{*}{\texttt{EM}} & BaryScore-SD-0.001$^{\Xi\varepsilon}$ & 77.78 & BaryScore-SD-0.001$^{\Xi\varepsilon}$ & 92.73 & BaryScore-SD-0.01$^{\Xi\varepsilon}$ & 90.01 \\
& BERTScore F1$^{\Xi\varepsilon}$ & 73.33 & BERTScore F1$^{\Xi\varepsilon}$ & 84.24 & BaryScore-W$^{\Xi\varepsilon}$ & 89.96 \\
& BaryScore-SD-0.01$^{\Xi\varepsilon}$ & 73.33 & BaryScore-SD-0.01$^{\Xi\varepsilon}$ & 84.24 & BERTScore F1$^{\Xi\varepsilon}$ & 88.67 \\
& MoverScore$^{\Xi\varepsilon}$ & 73.33 & MoverScore$^{\Xi\varepsilon}$ & 81.82 & SUPERT-Golden$^{\Xi\Delta}$ & 88.10 \\
& BaryScore-W$^{\Xi\varepsilon}$ & 68.89 & BaryScore-W$^{\Xi\varepsilon}$ & 80.61 & ROUGE-WE-3 F-Score$^{\Xi\varepsilon}$ & 87.93 \\
\midrule
\multirow{5}{*}{\texttt{SU}} & BaryScore-SD-0.001$^{\Xi\varepsilon}$ & 77.78 & BaryScore-SD-0.001$^{\Xi\varepsilon}$ & 90.30 & BARTScore-SH$^{\Xi\Delta}$ & 92.65 \\
& BaryScore-SD-5$^{\Xi\varepsilon}$ & 68.89 & BaryScore-SD-5$^{\Xi\varepsilon}$ & 83.03 & BERTScore Recall$^{\Xi\varepsilon}$ & 91.09 \\
& BaryScore-SD-10$^{\Xi\varepsilon}$ & 68.89 & BaryScore-SD-10$^{\Xi\varepsilon}$ & 83.03 & DepthScor$^{\Xi\varepsilon}$e & 90.71 \\
& BaryScore-SD-1$^{\Xi\varepsilon}$ & 64.44 & BaryScore-SD-1$^{\Xi\varepsilon}$ & 79.39 & SUPERT-Golden$^{\Xi\Delta}$ & 89.83 \\
& BaryScore-SD-0.5$^{\Xi\varepsilon}$ & 60.00 & BaryScore-SD-0.5$^{\Xi\varepsilon}$ & 76.97 & Compression$^{\text{¤§}}$ & 89.24 \\
\midrule
\multirow{5}{*}{\texttt{EG}} & BaryScore-SD-0.001$^{\Xi\varepsilon}$ & 77.78 & BaryScore-SD-0.001$^{\Xi\varepsilon}$ & 92.73 & DepthScore$^{\Xi\varepsilon}$ & 93.44 \\
& BaryScore-SD-5$^{\Xi\varepsilon}$ & 68.89 & BaryScore-SD-5$^{\Xi\varepsilon}$ & 78.18 & BARTScore-SH$^{\Xi\Delta}$ & 92.44 \\
& BaryScore-SD-10$^{\Xi\varepsilon}$ & 68.89 & BaryScore-SD-10$^{\Xi\varepsilon}$ & 78.18 & SUPERT-Golden$^{\Xi\Delta}$ & 92.21 \\
& BaryScore-SD-1$^{\Xi\varepsilon}$ & 64.44 & BaryScore-SD-1$^{\Xi\varepsilon}$ & 75.76 & MoverScore$^{\Xi\varepsilon}$ & 92.07 \\
& BaryScore-SD-0.5$^{\Xi\varepsilon}$ & 60.00 & BERTScore F1$^{\Xi\varepsilon}$ & 74.55 & BERTScore F1$^{\Xi\varepsilon}$ & 91.74 \\
\midrule
\multirow{5}{*}{\texttt{CX}} & BaryScore-SD-10$^{\Xi\varepsilon}$ & 76.41 & BaryScore-SD-10$^{\Xi\varepsilon}$ & 91.19 & DepthScore$^{\Xi\varepsilon}$ & 95.63 \\
& BaryScore-SD-5$^{\Xi\varepsilon}$ & 76.41 & BaryScore-SD-5$^{\Xi\varepsilon}$ & 91.19 & BERTScore Recall$^{\Xi\varepsilon}$ & 95.49 \\
& BaryScore-SD-1$^{\Xi\varepsilon}$ & 71.91 & BaryScore-SD-1$^{\Xi\varepsilon}$ & 87.54 & Compression$^{\text{¤§}}$ & 94.31 \\
& \textsc{chrF}$^{\Xi\text{§}}$ & 67.42 & Novelty-1$^{\text{¤§}}$ & 87.54 & BARTScore-SH$^{\Xi\Delta}$ & 93.83 \\
& Novelty-1$^{\text{¤§}}$ & 67.42 & BaryScore-SD-0.5$^{\Xi\varepsilon}$ & 85.11 & ROUGE-1 F-Score$^{\Xi\text{§}}$ & 93.35 \\
\bottomrule
\end{tabular}
\caption{Top 5 metrics per criterion per system-level correlation coefficient.}
\label{tab:top5_system}
\end{table}

\clearpage
\begin{minipage}{0.45\linewidth}
\section{Weighted macro F1-scores between automatic metrics and human criteria}
Here we report the weighted macro F1-scores between automatic metrics and human criteria obtained through the paired bootstrap resampling test.
\end{minipage}
\begin{figure}[h!]
    \centering
    \includegraphics[width=\textwidth]{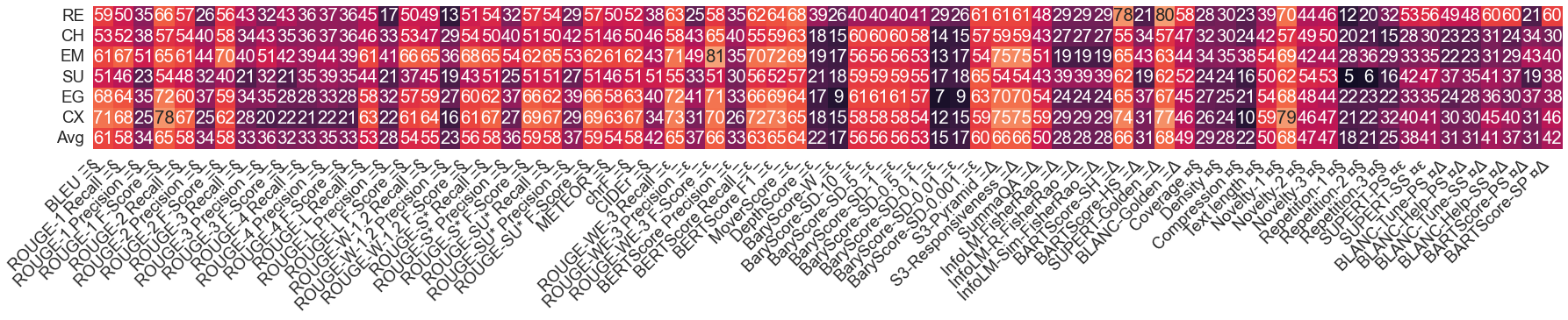}
    \caption{Weighted macro F1-scores of paired bootstrap
resampling}
    \label{fig:fscores2}
\end{figure}

\clearpage
\begin{minipage}{0.45\linewidth}
\section{Williams tests between automatic metrics}
Here we report the $p$-values of the Williams tests between automatic metrics for each criterion with story-level and system-level Pearson correlations.
\end{minipage}
\label{sec:williams}

\begin{figure}[h!]
    \centering
    \includegraphics[width=\textwidth]{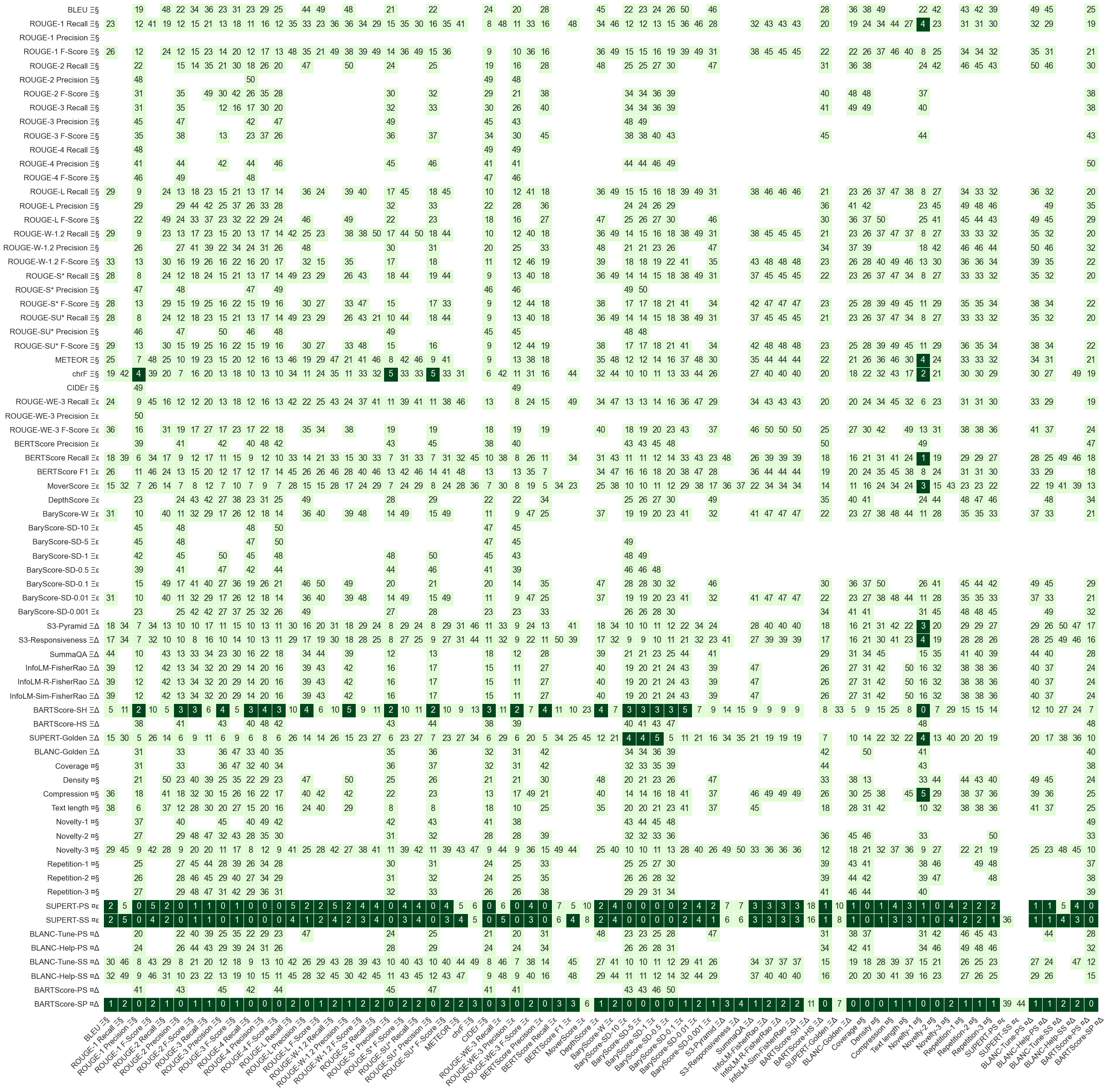}
    \caption{$p$-values (\%) of the Williams tests between automatic metrics for the \texttt{RE} criterion with story-level Pearson correlations. Green case means that the row metric has a higher correlation than the column metric, dark green means the increase is statistically significant ($p$ < 0.05).}
    \label{fig:williams_summary_pearson_Relevance}
\end{figure}

\begin{figure}[h]
    \centering
    \includegraphics[width=\textwidth]{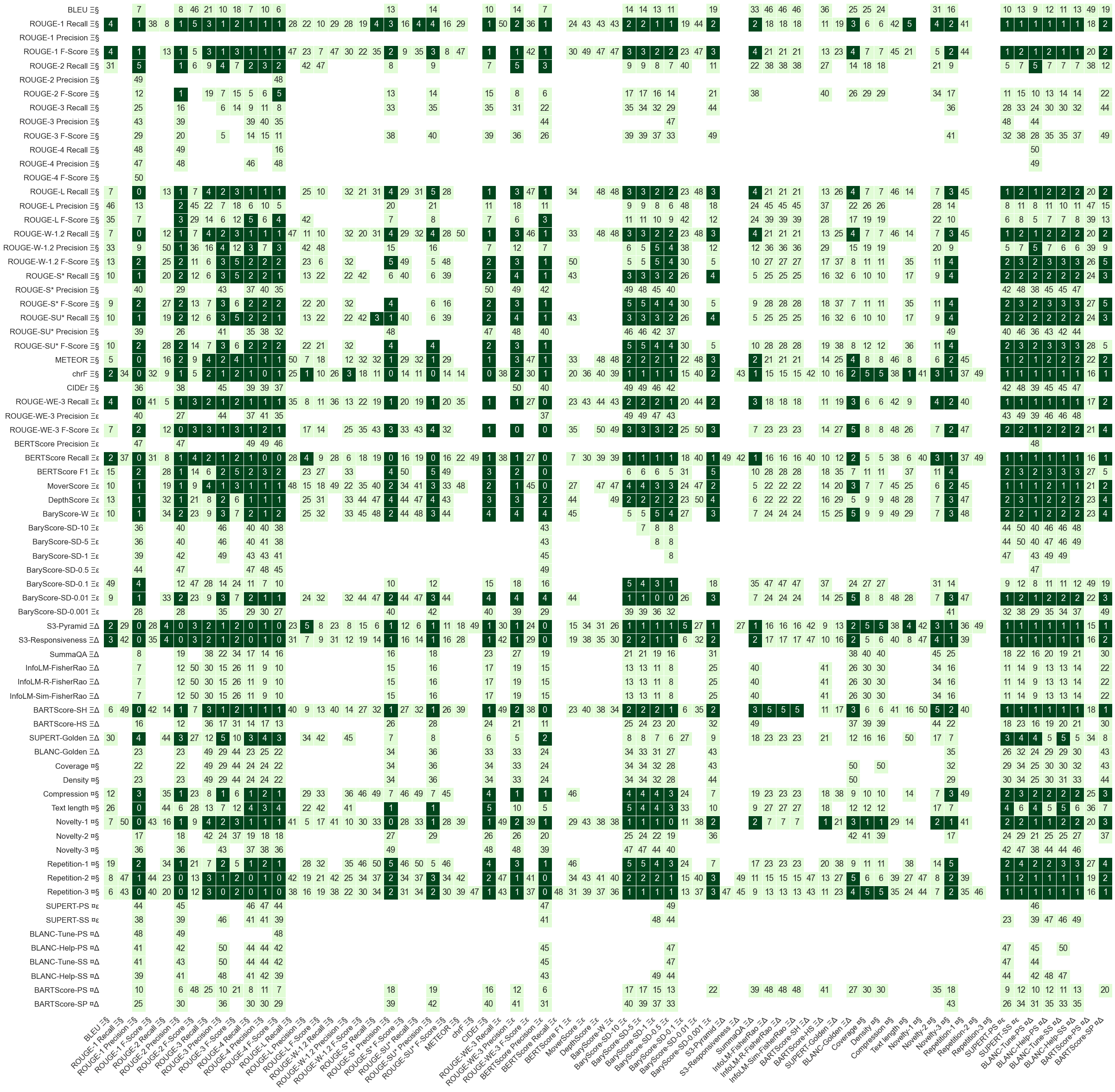}
    \caption{$p$-values (\%) of the Williams tests between automatic metrics for the \texttt{CH} criterion with story-level Pearson correlations. Green case means that the row metric has a higher correlation than the column metric, dark green means the increase is statistically significant ($p$ < 0.05).}
    \label{fig:williams_summary_pearson_Coherence}
\end{figure}

\begin{figure}[h]
    \centering
    \includegraphics[width=\textwidth]{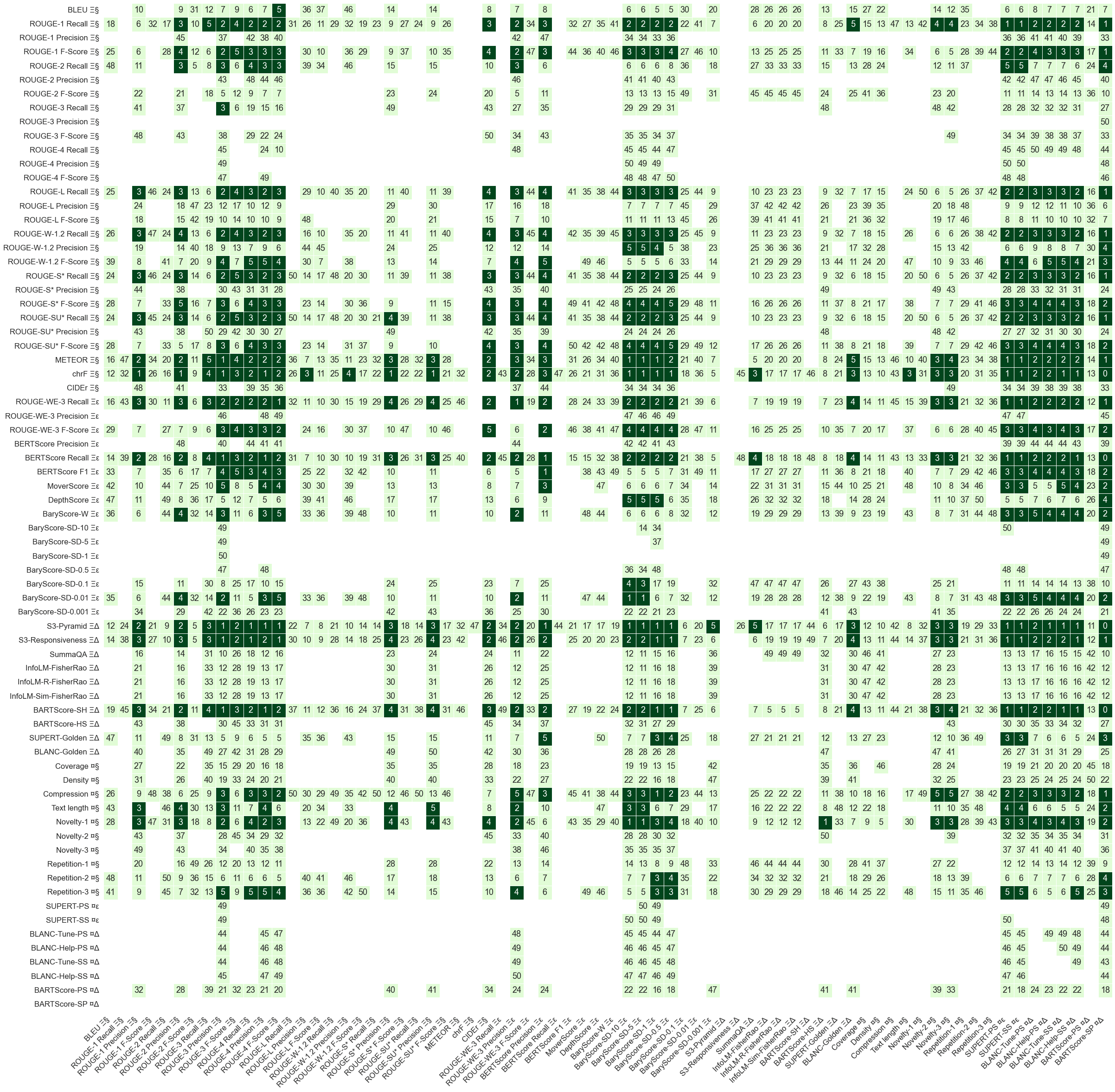}
    \caption{$p$-values (\%) of the Williams tests between automatic metrics for the \texttt{EM} criterion with story-level Pearson correlations. Green case means that the row metric has a higher correlation than the column metric, dark green means the increase is statistically significant ($p$ < 0.05).}
    \label{fig:williams_summary_pearson_Empathy}
\end{figure}

\begin{figure}[h]
    \centering
    \includegraphics[width=\textwidth]{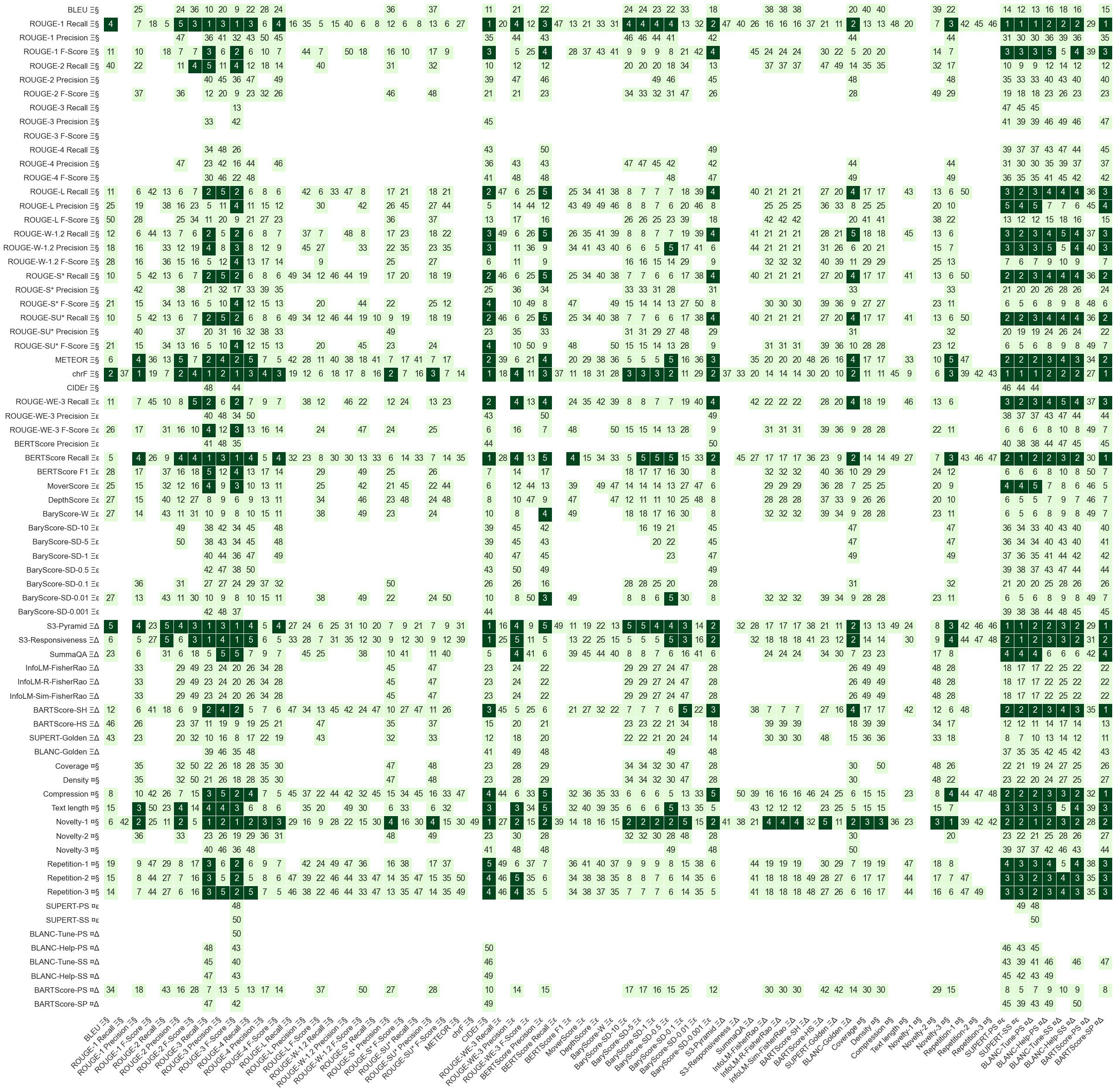}
    \caption{$p$-values (\%) of the Williams tests between automatic metrics for the \texttt{SU} criterion with story-level Pearson correlations. Green case means that the row metric has a higher correlation than the column metric, dark green means the increase is statistically significant ($p$ < 0.05).}
    \label{fig:williams_summary_pearson_Surprise}
\end{figure}

\begin{figure}[h]
    \centering
    \includegraphics[width=\textwidth]{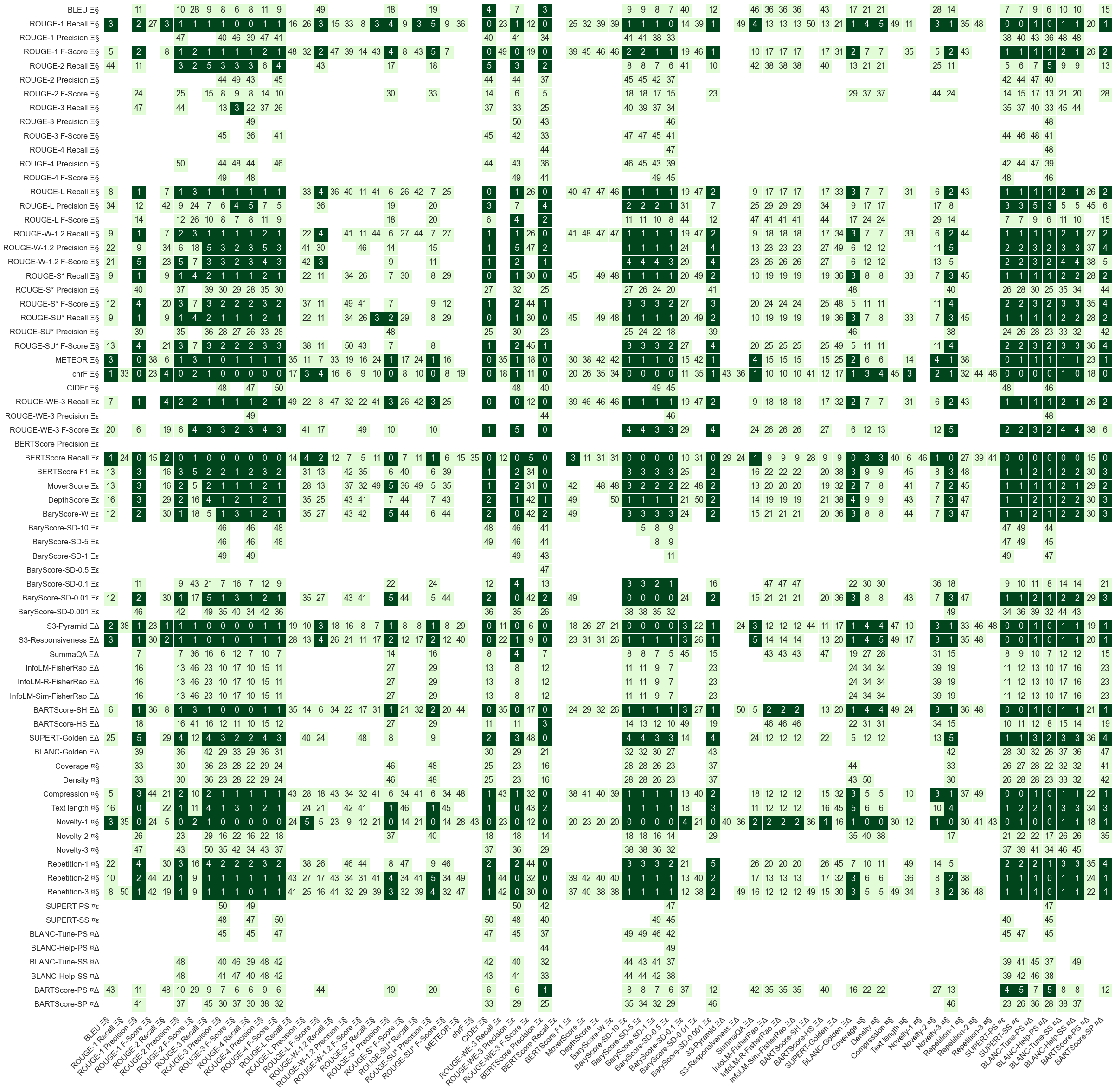}
    \caption{$p$-values (\%) of the Williams tests between automatic metrics for the \texttt{EG} criterion with story-level Pearson correlations. Green case means that the row metric has a higher correlation than the column metric, dark green means the increase is statistically significant ($p$ < 0.05).}
    \label{fig:williams_summary_pearson_Engagement}
\end{figure}

\begin{figure}[h]
    \centering
    \includegraphics[width=\textwidth]{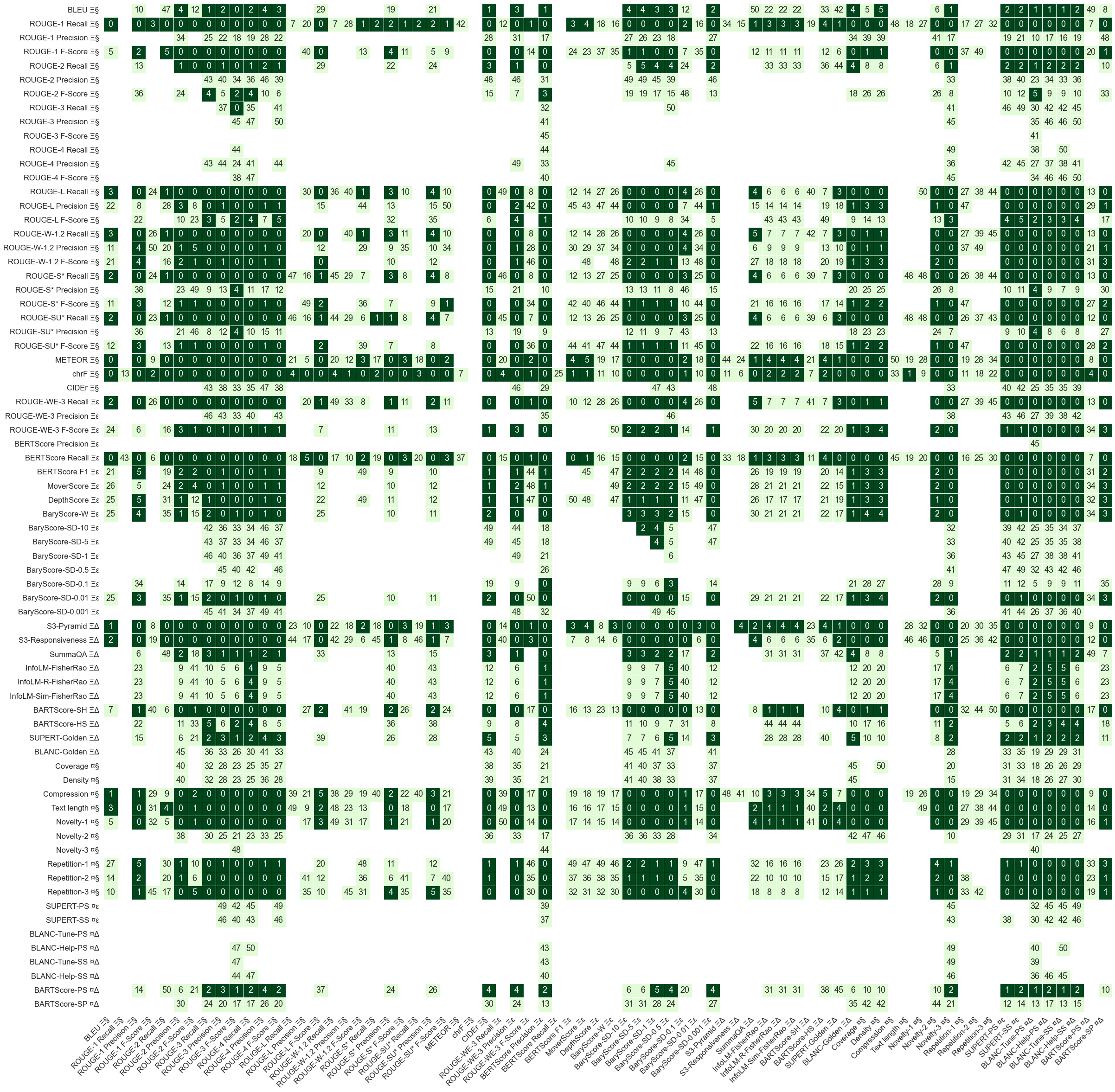}
    \caption{$p$-values (\%) of the Williams tests between automatic metrics for the \texttt{CX} criterion with story-level Pearson correlations. Green case means that the row metric has a higher correlation than the column metric, dark green means the increase is statistically significant ($p$ < 0.05).}
    \label{fig:williams_summary_pearson_Complexity}
\end{figure}

\begin{figure}[h]
    \centering
    \includegraphics[width=\textwidth]{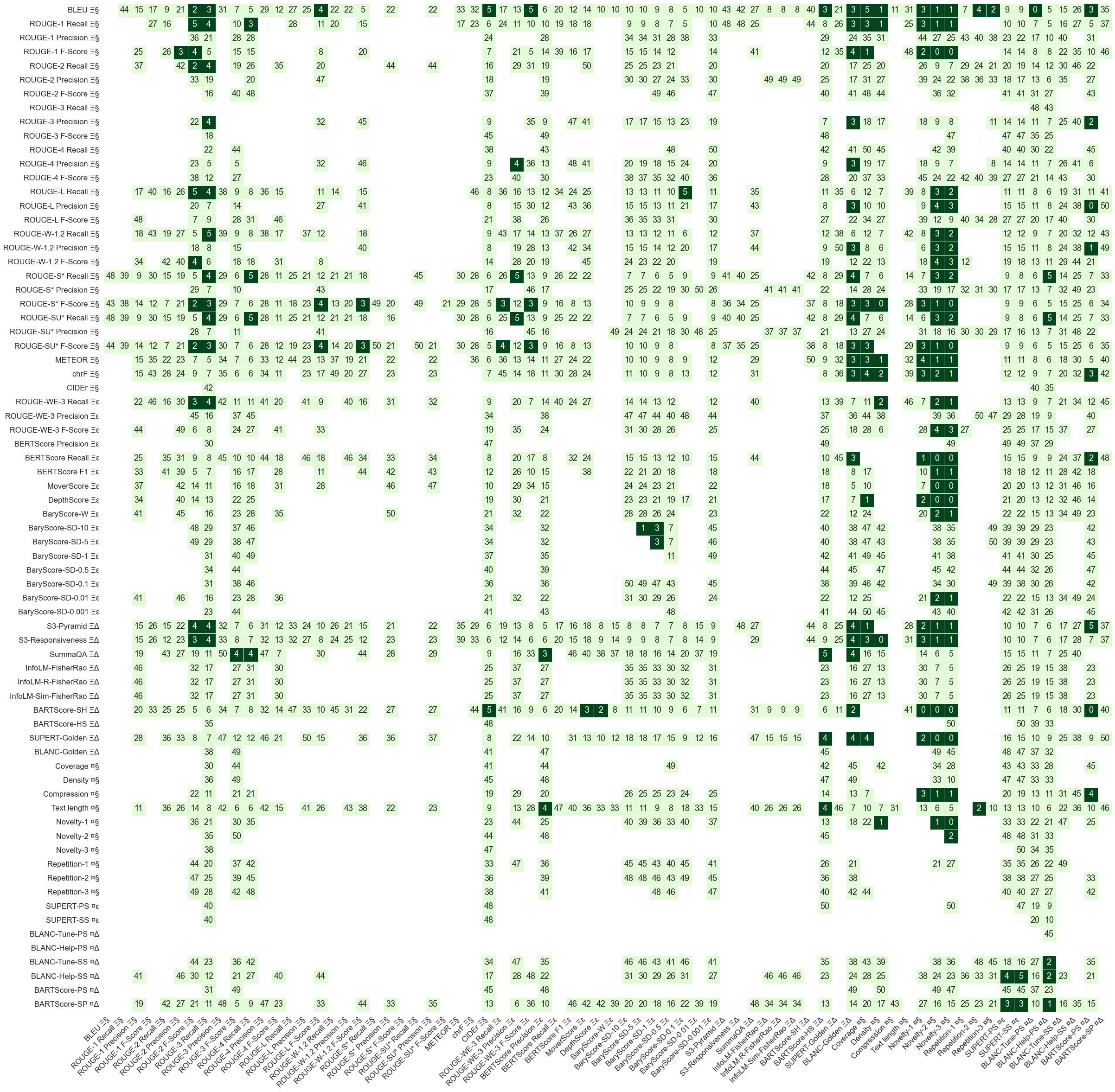}
    \caption{$p$-values (\%) of the Williams tests between automatic metrics for the \texttt{RE} criterion with system-level Pearson correlations. Green case means that the row metric has a higher correlation than the column metric, dark green means the increase is statistically significant ($p$ < 0.05).}
    \label{fig:williams_system_pearson_Relevance}
\end{figure}

\begin{figure}[h]
    \centering
    \includegraphics[width=\textwidth]{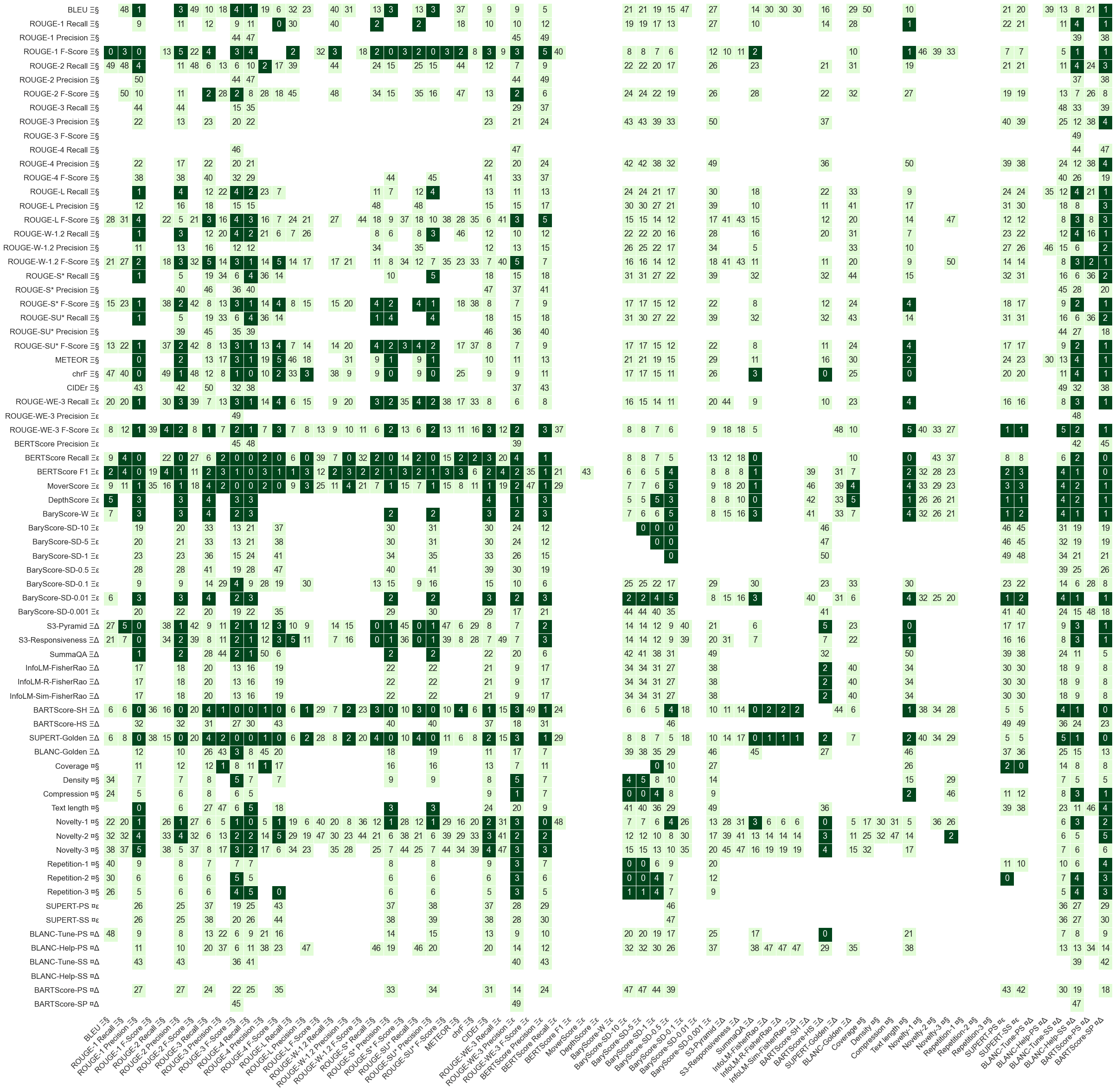}
    \caption{$p$-values (\%) of the Williams tests between automatic metrics for the \texttt{CH} criterion with system-level Pearson correlations. Green case means that the row metric has a higher correlation than the column metric, dark green means the increase is statistically significant ($p$ < 0.05).}
    \label{fig:williams_system_pearson_Coherence}
\end{figure}

\begin{figure}[h]
    \centering
    \includegraphics[width=\textwidth]{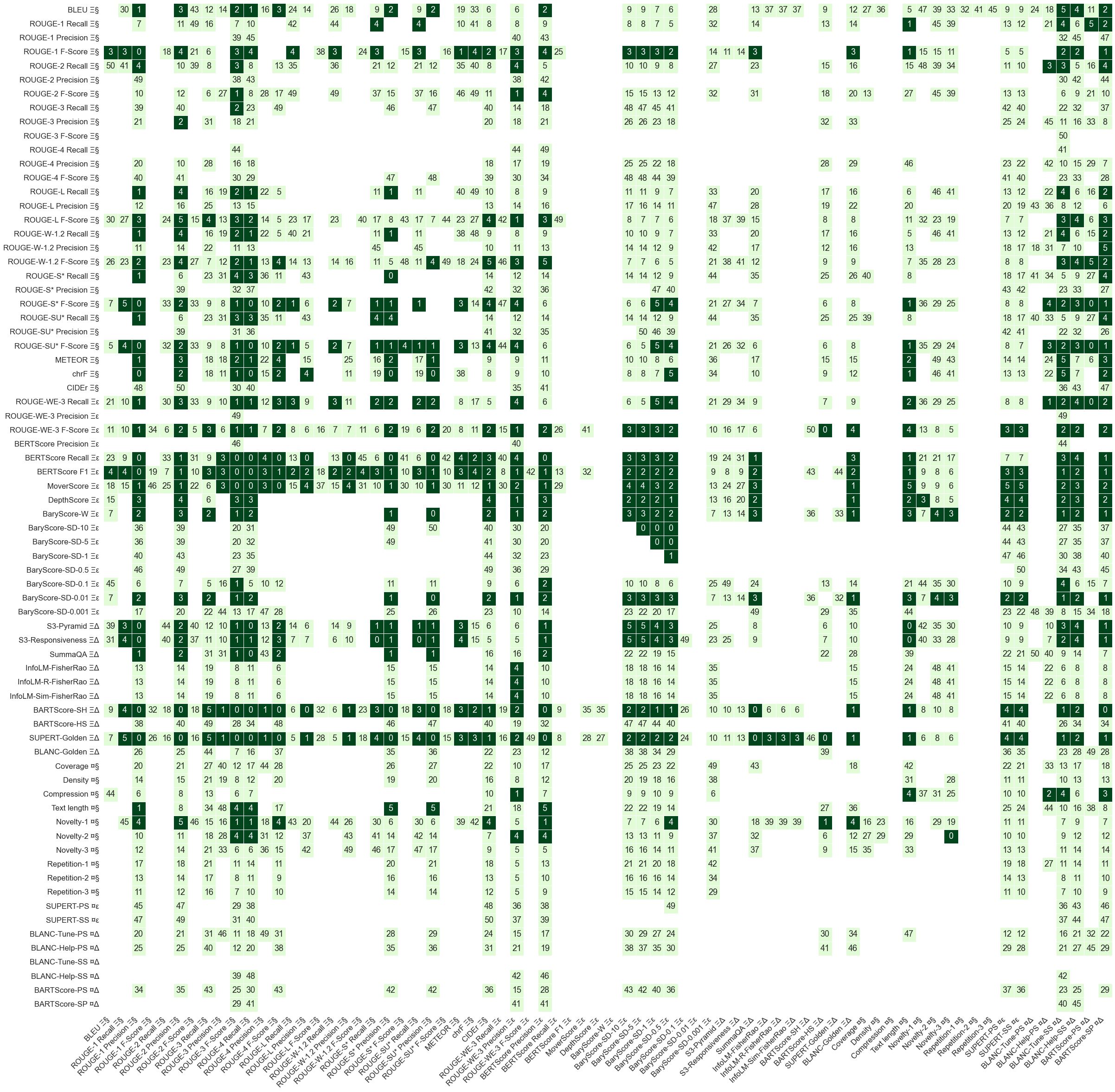}
    \caption{$p$-values (\%) of the Williams tests between automatic metrics for the \texttt{EM} criterion with system-level Pearson correlations. Green case means that the row metric has a higher correlation than the column metric, dark green means the increase is statistically significant ($p$ < 0.05).}
    \label{fig:williams_system_pearson_Empathy}
\end{figure}

\begin{figure}[h]
    \centering
    \includegraphics[width=\textwidth]{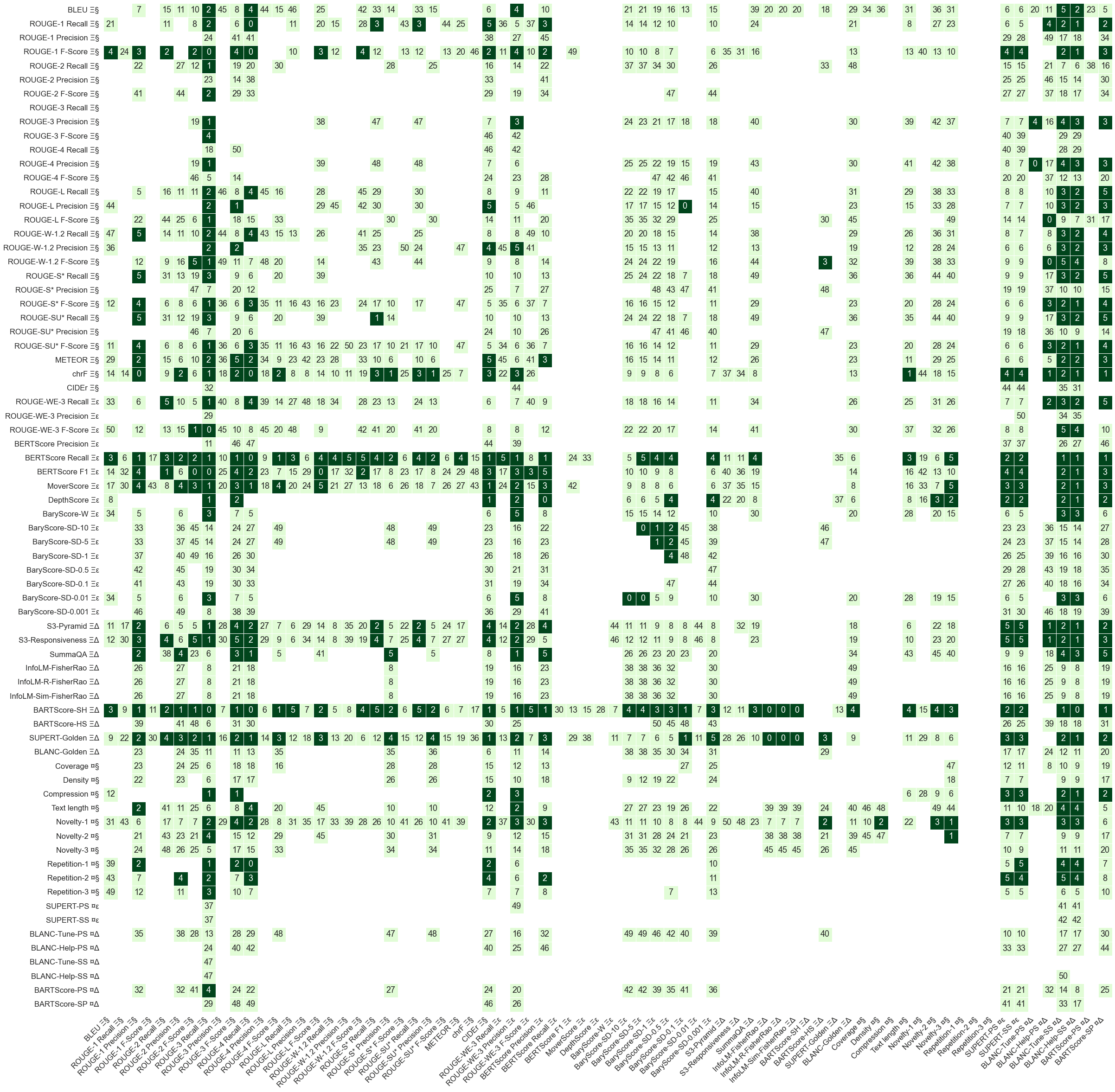}
    \caption{$p$-values (\%) of the Williams tests between automatic metrics for the \texttt{SU} criterion with system-level Pearson correlations. Green case means that the row metric has a higher correlation than the column metric, dark green means the increase is statistically significant ($p$ < 0.05).}
    \label{fig:williams_system_pearson_Surprise}
\end{figure}

\begin{figure}[h]
    \centering
    \includegraphics[width=\textwidth]{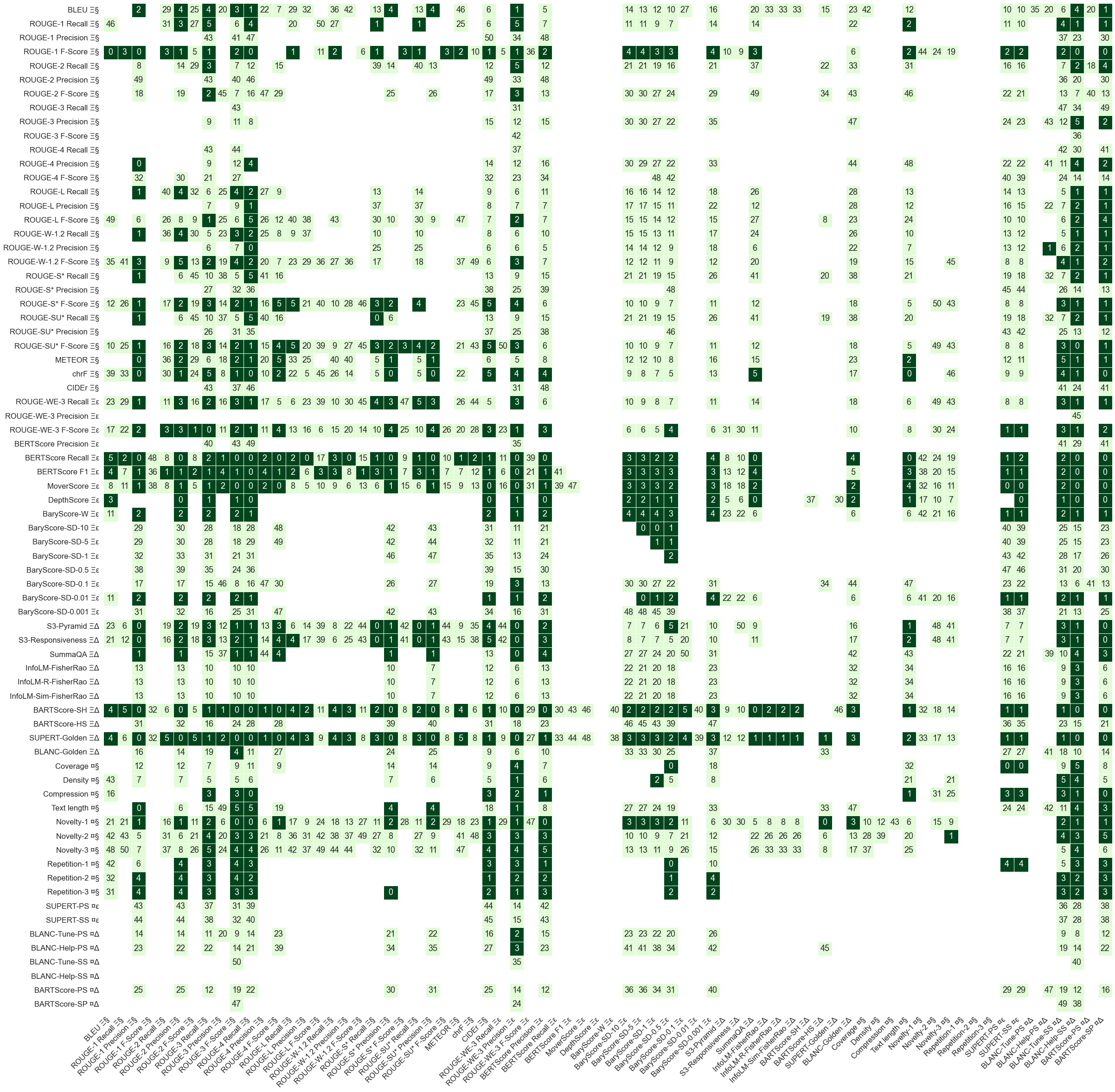}
    \caption{$p$-values (\%) of the Williams tests between automatic metrics for the \texttt{EG} criterion with system-level Pearson correlations. Green case means that the row metric has a higher correlation than the column metric, dark green means the increase is statistically significant ($p$ < 0.05).}
    \label{fig:williams_system_pearson_Engagement}
\end{figure}

\begin{figure}[h]
    \centering
    \includegraphics[width=\textwidth]{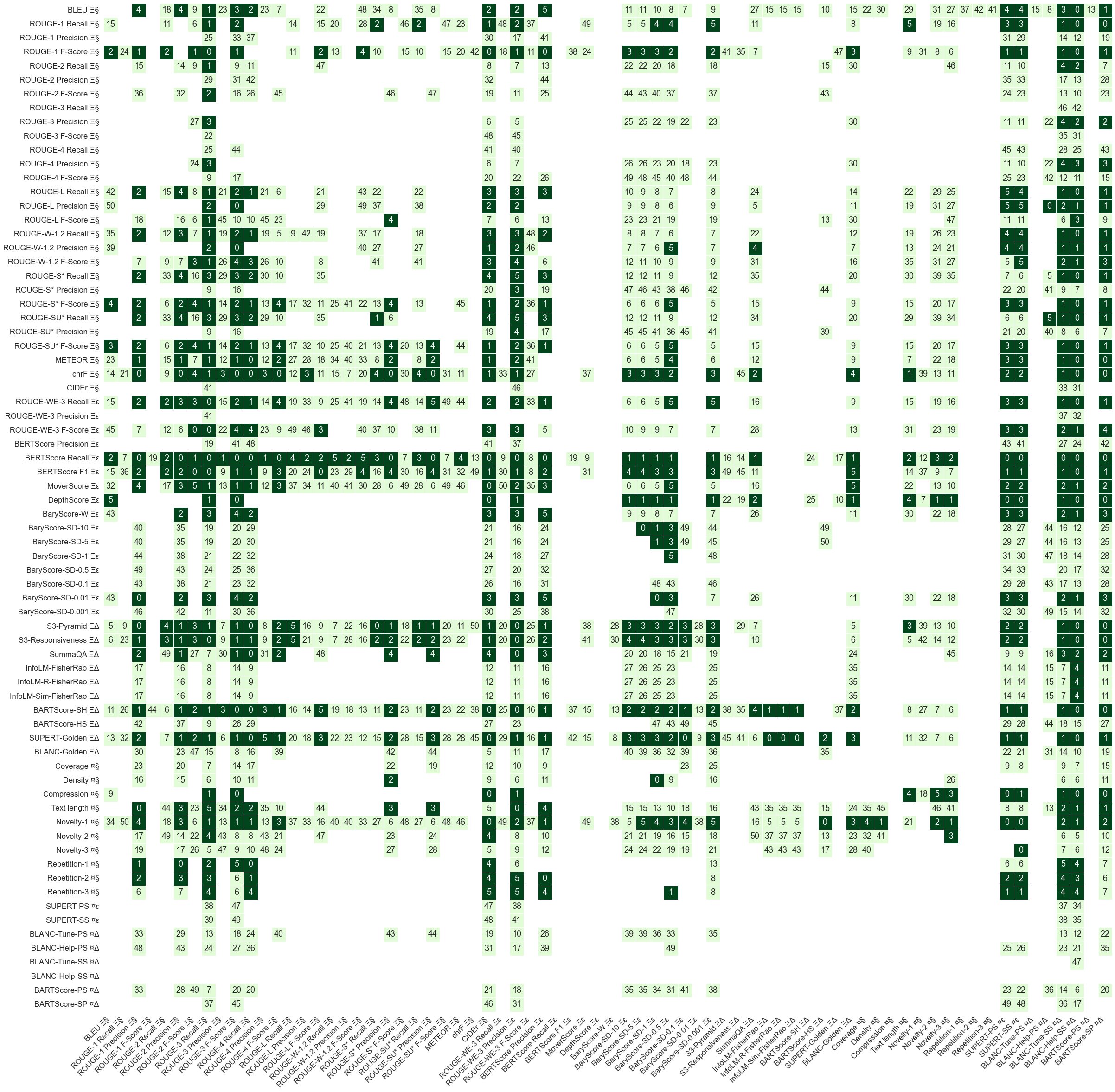}
    \caption{$p$-values (\%) of the Williams tests between automatic metrics for the \texttt{CX} criterion with system-level Pearson correlations. Green case means that the row metric has a higher correlation than the column metric, dark green means the increase is statistically significant ($p$ < 0.05).}
    \label{fig:williams_system_pearson_Complexity}
\end{figure}